\documentclass[final]{cvpr}

\usepackage{times}
\usepackage{epsfig}
\usepackage{graphicx}
\usepackage{amsmath}
\usepackage{amssymb}
\usepackage{multirow}
\usepackage{multicol}
\usepackage{booktabs}
\graphicspath{ {images/} }
\graphicspath{ {images_supp/} }
\usepackage{soul}
\usepackage{xcolor}

\usepackage{pifont}

\newcommand{\mypar}[1]{\vspace{1mm}\noindent{\bf #1}}

\definecolor{asparagus}{rgb}{0.53, 0.66, 0.42}
\definecolor{armygreen}{rgb}{0.29, 0.33, 0.13}
\definecolor{awesome}{rgb}{1.0, 0.13, 0.32}
\definecolor{applegreen}{rgb}{0.55, 0.71, 0.0}

\DeclareMathOperator*{\argmax}{argmax}

\usepackage[pagebackref=true,breaklinks=true,colorlinks,bookmarks=false]{hyperref}

\def\cvprPaperID{} 
\def\confYear{CVPR 2021}

\begin{document}

\title{RGB-D Local Implicit Function for Depth Completion of Transparent Objects}

\author{
Luyang Zhu\textsuperscript{1,2}\footnotemark\qquad 
Arsalan Mousavian\textsuperscript{2}\qquad 
Yu Xiang\textsuperscript{2}\qquad
Hammad Mazhar\textsuperscript{2}\qquad \\
Jozef van Eenbergen\textsuperscript{2}\qquad
Shoubhik Debnath\textsuperscript{2}\qquad
Dieter Fox\textsuperscript{1,2} \\
\textsuperscript{1}University of Washington\qquad 
\textsuperscript{2}NVIDIA
}

\maketitle

\footnotetext[1]{Work done while author was an intern at NVIDIA.}


\begin{abstract}
Majority of the perception methods in robotics require depth information provided by RGB-D cameras. However, standard 3D sensors fail to capture depth of transparent objects due to refraction and absorption of light. In this paper, we introduce a new approach for depth completion of transparent objects from a single RGB-D image. Key to our approach is a local implicit neural representation built on ray-voxel pairs that allows our method to generalize to unseen objects and achieve fast inference speed. Based on this representation, we present a novel framework that can complete missing depth given noisy RGB-D input. We further improve the depth estimation iteratively using a self-correcting refinement model. To train the whole pipeline, we build a large scale synthetic dataset with transparent objects. Experiments demonstrate that our method performs significantly better than the current state-of-the-art methods on both synthetic and real world data. In addition, our approach improves the inference speed by a factor of ~\textbf{20} compared to the previous best method, ClearGrasp~\cite{sajjan2020clear}. Code and dataset will be released at~\url{https://research.nvidia.com/publication/2021-03_RGB-D-Local-Implicit}.
\end{abstract}

`

\section{Introduction}
\label{sec::intro}
Depth data captured from RGB-D cameras has been widely used in many applications such as augmented reality and robot manipulation. Despite their popularity, commodity-level depth sensors, such as structured-light cameras and time-of-flight cameras, fail to produce correct depth for transparent objects due to the lack of light reflection. As a result, many algorithms utilizing RGB-D data cannot be directly applied to recognize transparent objects which are very common in household scenarios. 

Previous works on estimating geometry of transparent objects are often studied under controlled settings~\cite{FRT18,yeung2011matting}. Recently, Li \textit{et al.}~\cite{Li_2020_CVPR} proposed a physically-based neural network to reconstruct 3D shape of transparent objects from multi-view images. Although their method is less restricted compared to previous ones, it still requires the environment map and the refractive index of transparent objects.  ClearGrasp~\cite{sajjan2020clear} achieves impressive results on depth completion of transparent objects. It first predicts masks, occlusion boundaries and surface normals from RGB images using deep networks, and then optimizes initial depth based on the network predictions. However, the optimization requires transparent objects to have contact edges with other non-transparent objects. Otherwise, the depth in transparent area becomes undetermined and can be assigned random value. In addition, it can not be deployed in real time applications due to the expensive optimization process.

\begin{figure}
\begin{center}
   \includegraphics[width=1.0\linewidth]{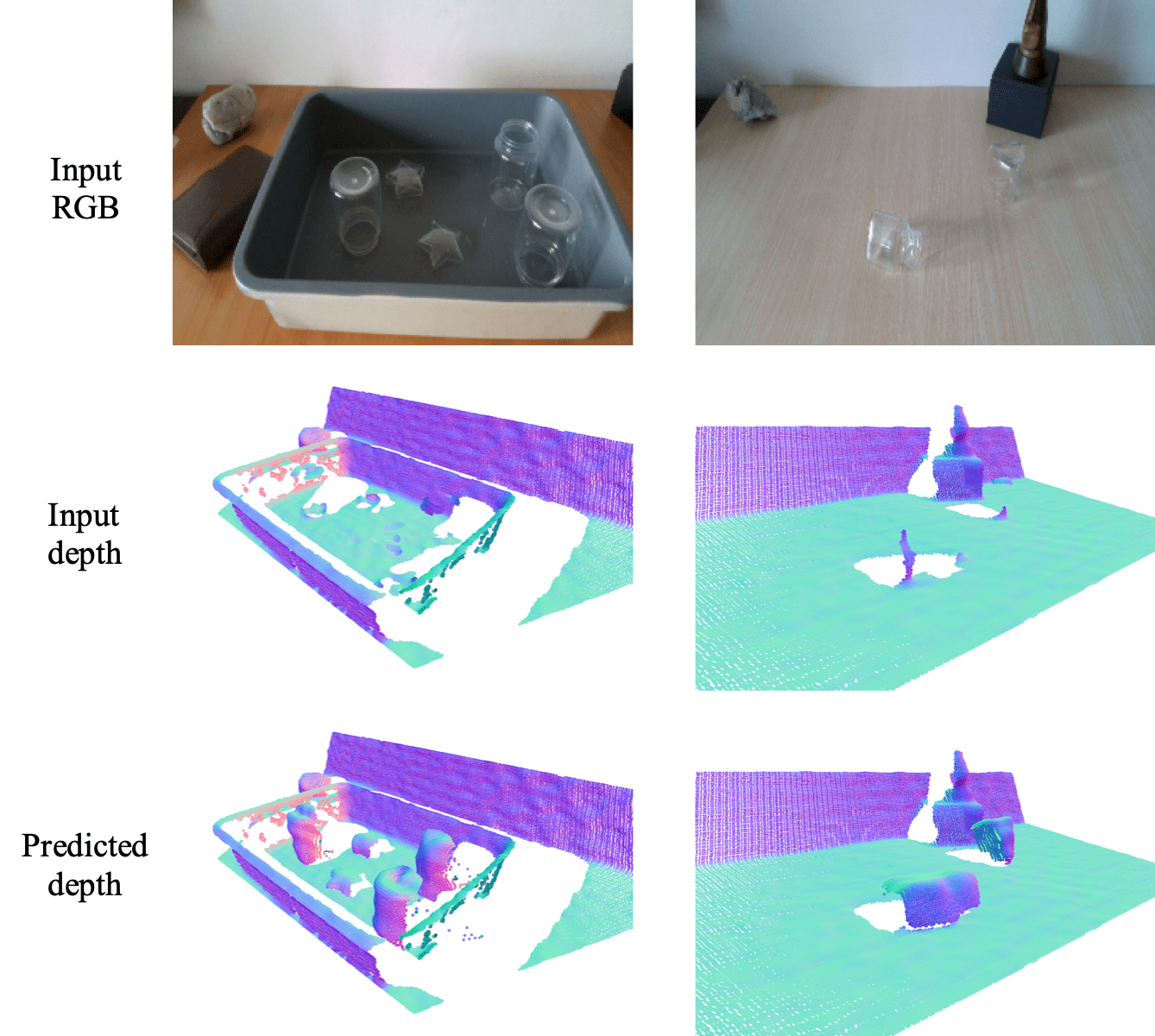}
\end{center}
 \vspace{-3mm}
   \caption{Our method can predict the depth of unseen transparent objects from a noisy RGB-D image. 
   We back-project the depth map into the point cloud and render it in a novel viewpoint to better visualize the 3D shape. Zoom in to see details.}
\vspace{-3mm}
\label{fig:lidf}
\end{figure}

In this paper, to overcome the limitations of existing works, we present a fast end-to-end framework for depth completion of transparent objects from a single RGB-D image. The core to our approach is a Local Implicit Depth Function (LIDF) defined on ray-voxel pairs consisting of camera rays and their intersecting voxels. The motivations for LIDF are: 1) The depth of a transparent object can be inferred from its color and the depth of its non-transparent neighborhood. In particular, color can provide useful visual cues for the 3D shape and curvature while local depth helps to reason about the spatial arrangement and location of transparent objects. 2) Directly regressing the complete depth map using a deep network can easily overfit to the objects and scenes in the training data. By learning at the local scale (a voxel in our case) instead of the whole scene, LIDF can generalize to unseen objects because different objects may share similar local structures. 3) Voxel grids provide a natural partition of the 3D space. By defining implicit function on ray-voxel pairs, we can significantly reduce the inference time as the model only needs to consider occupied voxels intersected by the camera ray. Based on these motivations, we present a model to estimate the depth of a pixel by learning the relationship between the camera ray and its intersecting voxels given the color and local depth information.
To further utilize the geometry of transparent object itself, we propose a depth refinement model to update the prediction iteratively by combining the input RGB, input depth points and the predicted depth from LIDF. To train the whole pipeline, we create a large scale synthetic dataset, Omiverse Object dataset, using the NVIDIA Omniverse platform~\cite{omniverse}. Our dataset provides over 60,000 images including both transparent and opaque objects in different scenes. The dataset is generated with diverse object models and poses, lighting conditions, camera viewpoints and background textures to close the sim-to-real gap. Experiments show that training on the Omniverse Object dataset can boost the performance for both our approach and competing methods in real-world testing cases.

Our approach is inspired by recent advances in neural radience field~\cite{mildenhall2020nerf,schwarz2020graf,liu2020neural}. NeRF~\cite{mildenhall2020nerf} can learn a continuous function of the 3D geometry and appearance of a scene, thus achieving accurate reconstruction and rendering results. However, NeRF has slow inference speed due to inefficient 3D points sampling. Our approach tackles this problem by querying intersecting voxels instead of 3D points along the ray. 
NSVF~\cite{liu2020neural} also utilizes a sparse voxel grid to reduce the inference time, but it still needs to sample 3D points inside the voxel while our method directly learns the offset of a ray-voxel pair to obtain the possible terminating 3D location of the ray. Experiments demonstrate that learning offsets can produce better depth than sampling 3D points based on heuristic strategies. In addition, NeRF-based methods needs to train a network for every new scene to model the complex geometry and appearance whereas our method generalizes to unseen objects and scenes in depth completion of transparent objects.

Our contributions are summarized as follows: 1) We propose LIDF, a novel implicit representation defined on ray-voxel pairs, leading to fast inference speed and good generality. 2) We present a two-stage system, including networks to learn LIDF and a self-correcting refinement model, for depth completion of transparent objects. 3) We build a large scale synthetic dataset proved to be useful to transparent objects learning. 4) Our full pipeline is evaluated qualitatively and quantitatively, and outperform the current state-of-the-art in terms of accuracy and speed.

\begin{figure*}
\begin{center}
  \includegraphics[width=1.0\linewidth]{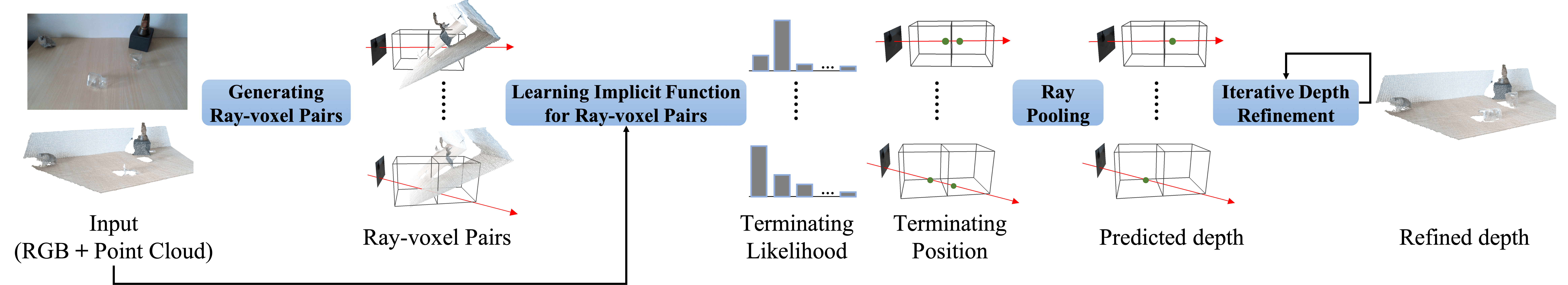}
\end{center}
\vspace{-2mm}
  \caption{Pipeline overview. Our method takes as input RGB-D images with missing depth for transparent objects, and predicts the full depth map. We first generate ray-voxel pairs by finding all occupied voxels intersected by a ray. Then we learn a local implicit function for each pair to estimate the terminating probability and position of the ray inside the voxel. The prediction of all pairs along the ray are accumulated by our ray pooling module to get the initial depth prediction. Finally, we improve the result by iterative depth refinement.}
\vspace{-3mm}
\label{fig:pipeline}
\end{figure*}

\section{Related Work}
\label{sec::related}
\mypar{Depth estimation.} Depth estimation can be classified into three categories based on the input. Several methods have been proposed to directly regress the depth map from the color image using convolutional neural networks~\cite{eigen2014depth,Ranftl2020,laina2016deeper,chen2016single,fu2018deep,garg2016unsupervised,godard2017unsupervised}. Most of them are trained on large scale datasets generated from RGB-D cameras, thus they can only reproduce the raw depth scan. Our method, on the contrary, focuses on the depth estimation for transparent objects where depth sensor typically fails. Another line of related work explores the task of depth completion given RGB images and sparse sets of depth measurements~\cite{mal2018sparse,chen2019learning,qiu2019deeplidar,xu2019depth,cheng2018depth,park2020non}. 
These works improve the depth estimation over color-only methods, but they still produce low quality results because of limited information provided by sparse depth. 
Our method falls into the third category which tries to complete depth maps given noisy RGB-D images. Barron and Malik~\cite{barron2013intrinsic} propose a joint optimization for intrinsic images. Firman \textit{et al.}~\cite{firman2016structured} predict unobserved voxels from a single depth image using the voxlet representation. Matsuo and Aoki~\cite{matsuo2015depth} reconstruct depths by ray-tracing to estimated local tangents. Recent works~\cite{zhang2018deep,sajjan2020clear} estimate surface normals and occlusion boundaries only from color images using deep networks and solve a global optimization based on those predictions as well as observed depths. The optimization is very slow and produces bad results if the network predictions are not accurate. We address these limitations by learning a implicit function using color and local depth jointly. Experiment shows that our method can achieve better results and $20\times$ speedup compared to~\cite{sajjan2020clear}.

\mypar{Transparent objects.} Transparent objects have been studied in various computer vision tasks, including 
object pose estimation~\cite{klank2011transparent,lysenkov2013pose,lysenkov2013recognition,phillips2016seeing,Liu_2020_CVPR}, 3D shape reconstruction~\cite{albrecht2013seeing,han2015fixed,qian20163d,Li_2020_CVPR,sajjan2020clear}
and segmentation~\cite{Kalra_2020_CVPR}. However, most of these works assume known background patterns~\cite{han2015fixed,qian20163d}, known object 3D models~\cite{klank2011transparent,lysenkov2013pose,phillips2016seeing}, or multi view/stereo input~\cite{Liu_2020_CVPR,Li_2020_CVPR}. Our approach does not require any priors and can estimate the depth of transparent objects from a single view RGB-D image. Sajjan \textit{et al.}~\cite{sajjan2020clear} is the closest work to ours. However, they pretrain their networks on out-of-domain real datasets while our method is trained purely on synthetic data but achieves better performance.

\mypar{Implicit function learning.} Our method takes inspiration from the implicit neural models~\cite{chen2019learningimplicit,mescheder2019occupancy,park2019deepsdf,genova2019learning,saito2019pifu,xu2019disn,chibane2020implicit}, which are not restricted by topology and can represent the 3D surface continuously. Most of these methods~\cite{chen2019learningimplicit,mescheder2019occupancy,park2019deepsdf} can not scale to scene level as they only use a single latent vector to encode the shape. Voxel-based implicit representations~\cite{chabra2020deep,jiang2020lig,Peng2020ECCV} address this limitation by learning local shape priors. However, they require coarse voxel grids or sparse point clouds as input, which already provide some information about 3D shapes. Our task is more challenging as the point cloud of transparent objects is totally missing. 

Recently, Mildenhall \textit{et al.} ~\cite{mildenhall2020nerf} propose to represent scenes as neural radiance fields, achieving impressive results on novel view synthesis. However, their method needs to be retrained for every new scene and takes a long time to render one image. Schwarz \textit{et al.}~\cite{schwarz2020graf} solves the first drawback by learning a generative radiance field, but their results are limited to single object or human face. Liu \textit{et al.}~\cite{liu2020neural} tackles the second problem by introducing sparse voxel grids, but they still need to sample points based on heuristic strategies while our method learns to estimate the possible terminating position for ray-voxel pairs. 


\section{Method}
\label{sec::method}

Figure~\ref{fig:pipeline} provides the overview of our proposed end-to-end system for depth completion of transparent objects. Our approach takes as input a RGB image $\mathcal{I} \in \mathbb{R}^{H \times W \times 3}$, an incomplete depth image $\mathcal{D} \in \mathbb{R}^{H\times W}$ and camera intrinsics $K \in \mathbb{R}^{3 \times 3}$, and predicts the full depth map $\Hat{\mathcal{D}} \in \mathbb{R}^{H\times W}$, where $H$ and $W$ are height and width of the input image.

Our method consists of two stages: The first stage learns LIDF defined on ray-voxel pairs to represent the local geometry of the scene, which is done in three steps. First, ray-voxel pairs are generated by marching a camera ray through the pixel to find occupied voxels intersected by the ray (Section~\ref{subsec::gen_pair}). Then networks are trained end-to-end to learn the terminating probability and position of the ray inside the voxel given RGB embedding, voxel embedding and ray embedding (Section~\ref{subsec::IF_pair}). Finally, predictions of all pairs along the ray are accumulated by a ray pooling module to get the depth estimation (Section~\ref{subsec::ray_pool}). 
Voxel embedding in the first stage does not encode the geometry of transpaernt objects due to the missing depth. The second stage of our method (Section~\ref{subsec::drm}) presents a refinement model to solve this problem. The refinement model first recomputes the voxel and ray embeddings based on input and predicted depth points. After that, it refines the depth estimation by combining the new voxel and ray embeddings with the original RGB embedding. The formulation of the refinement model allows for iteratively applying fine adjustments that preserve common 3D structures such as shapes of the objects, how the objects are lying on the table, and objects following the direction of gravity. Experiments demonstrate the generalization power of the depth refinement model on unseen objects in the real world.

\subsection{Generating Ray-voxel Pairs}
\label{subsec::gen_pair}
\begin{figure}
\begin{center}
   \includegraphics[width=1.0\linewidth]{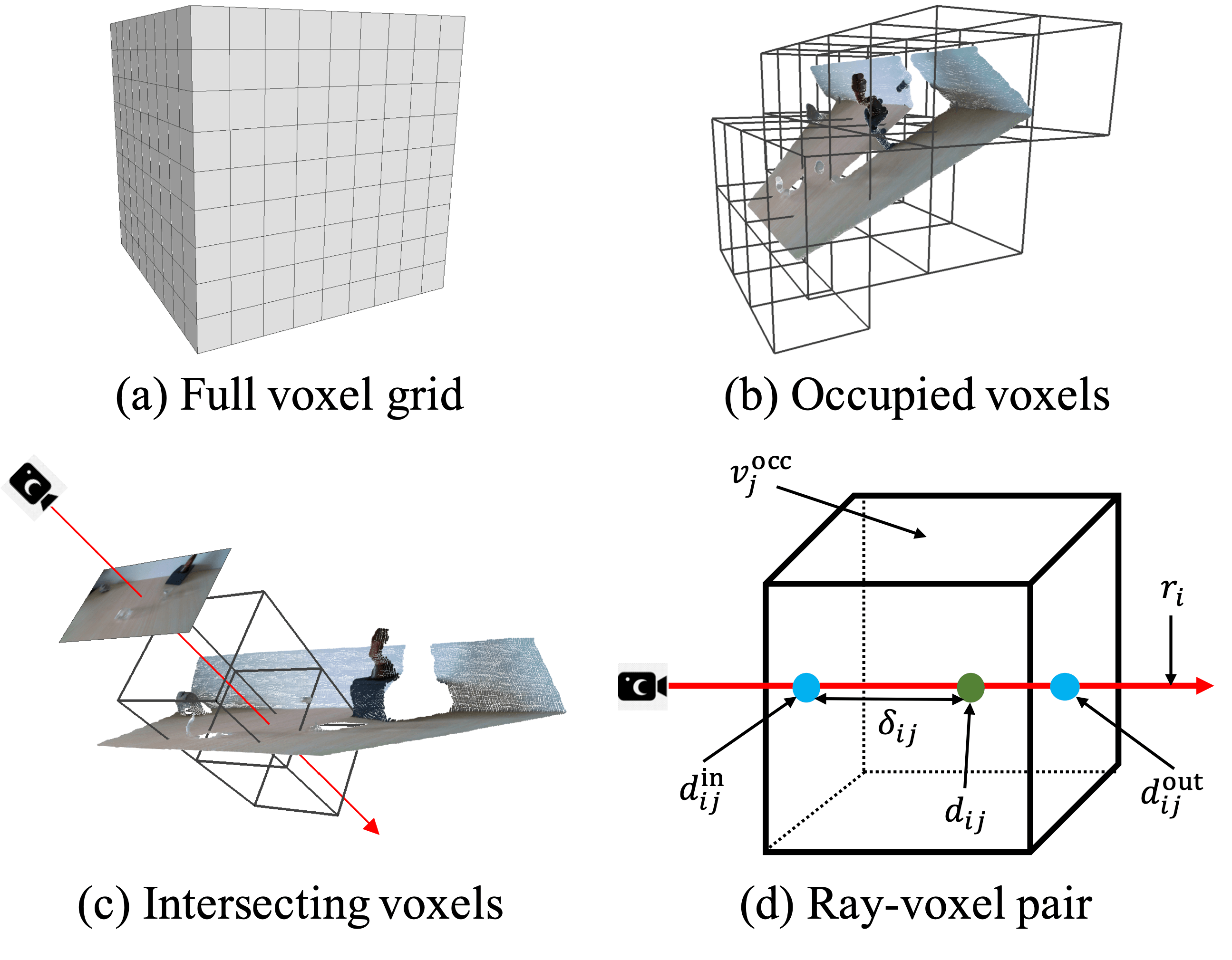}
\end{center}
\vspace{-5mm}
   \caption{Ray-voxel pair generation. (a) Predefined voxel grid. (b) Occupied voxels. (c) Intersecting voxels along the ray. (d) Variables defined for the ray-voxel pair.} 
\label{fig:gen_pair}
\end{figure}

To predict the missing depth of a pixel, we first need to find all the occupied voxels intersected by its camera ray.  Figure~\ref{fig:gen_pair} provides different steps for the ray-voxel pair generation. We start by dividing a fixed workspace into an axis-aligned voxel grid $\mathbf{V}$ in camera coordinates, where our system only completes the depth for any object inside.
After that, the input depth map $\mathcal{D}$ is back-projected into an “organized” point cloud $P \in \mathbb{R}^{H\times W\times 3}$ using the camera intrinsics $K$. We filter out points with zero depth to get the valid point cloud $P^\text{valid} \subset P$. A voxel is considered as occupied if it has at least one valid depth point. Based on this definition, the set of occupied voxels $\mathbf{V}^\text{occ}$ can be computed by doing a boundary test between $P^\text{valid}$ and all voxel cells $v \in \mathbf{V}$. Finally, we shoot a camera ray $r_i$ through the $i$-th pixel $\mathcal{I}_i$ and apply the ray-AABB intersection test~\cite{kay1986ray} between $r_i$ and $\mathbf{V}^\text{occ}$ using a customized CUDA kernel which allows for real time computation. 

The generated ray-voxel pairs (Figure~\ref{fig:gen_pair} (d)) are denoted by $\Phi_{ij} = (r_i, v^\text{occ}_{j}, d^\text{in}_{ij}, d^\text{out}_{ij})$, where $r_i$ is the ray direction for the $i$-th pixel $\mathcal{I}_i$, $v^\text{occ}_{j}$ is the voxel intersected by the ray $r_i$ with index $j$ in $\mathbf{V}^\text{occ}$, $d^\text{in}_{ij}$ and $d^\text{out}_{ij}$ are the position of the entering and leaving intersection point respectively.

\subsection{Learning LIDF for Ray-voxel Pairs} 
\label{subsec::IF_pair}
LIDF is defined on ray-voxel pairs to reason about the relationship between camera rays and their intersecting voxels. Specifically, we want to estimate the terminating probability and position of the camera ray inside the intersecting voxel given the following inputs: 1) RGB embedding corresponding to the pixel of the camera ray; 2) Voxel embedding computed from valid points inside the intersecting voxel; 3) Ray embedding which encodes the high frequency information of the ray direction and ray-voxel intersection points. As illustrated in Figure~\ref{fig:lidf}, the learning process of LIDF for the ray-voxel pair $\Phi_{ij}$ can be formulated as:
\begin{align}
& p^\text{end}_{ij} = \mathcal{F}_\text{prob}(\mathcal{H}_{ij}) \\
& d_{ij} = d^\text{in}_{ij} + \delta_{ij} \cdot r_i, \delta_{ij} = \mathcal{F}_\text{pos}(\mathcal{H}_{ij}) \\
& \mathcal{H}_{ij} = \mathcal{H}^\text{rgb}_{i} \oplus \mathcal{H}^\text{vox}_{j} \oplus \gamma(r_i) \oplus \gamma(d^{in}_{ij}) \oplus \gamma(d^\text{out}_{ij}),
\end{align}
where $p^\text{end}_{ij}$ is the probability of the ray $r_i$ terminating in the voxel $v^\text{occ}_{j}$; $d_{ij}$ is the terminating position inside $v^\text{occ}_{j}$ for $r_i$; $\delta_{ij}$ is the distance from $d^\text{in}_{ij}$ to $d_{ij}$ along the ray; 
$\mathcal{H}_{ij}$ is the input feature embedding for $\Phi_{ij}$; $\mathcal{H}^\text{rgb}_{i}$ is the RGB embedding for pixel $\mathcal{I}_i$; $\mathcal{H}^\text{vox}_{j}$ is the voxel embedding for $v^\text{occ}_{j}$; $\gamma(\cdot)$ is the high frequency positional encoding function~\cite{mildenhall2020nerf}; $\oplus$ is the concatenation operator for feature vectors. $\mathcal{F}_\text{prob}$ is parametrized as a MLP while $\mathcal{F}_\text{pos}$ is parametrized as a separate MLP with iterative error feedback~\cite{carreira2016human}. 

\mypar{RGB Embedding.} The RGB embedding $\mathcal{H}^\text{rgb}$ encodes the color information of all pixels. Given the input image $\mathcal{I}$, we first apply ResNet34-8s network~\cite{xiang2020learning} $\Psi_\text{rgb}$ to get a dense feature map $\Psi_\text{rgb}(\mathcal{I}) \in \mathbb{R}^{H \times W \times C_r}$, where $H$ is the image height, $W$ is the image width and $C_r$ is the dimension of the dense feature map. To compute the RGB Embedding $\mathcal{H}^\text{rgb}_{i}$ for the i-th pixel $\mathcal{I}_i$, we use ROIAlign~\cite{he2017mask} to pool features from a local patch of $\Psi_\text{rgb}(\mathcal{I})$ centered at $\mathcal{I}_i$. We set $C_r=32$ in our experiments. For ROIAlign, the size of input ROI is $8 \times 8$ and the size of output feature map is $ 2 \times 2$. We directly flatten the output feature map of ROIAlign to get the final RGB embedding.

\mypar{Voxel Embedding.} The voxel embedding $\mathcal{H}^\text{vox}$ encodes the geometry information for occupied voxels $\mathbf{V}^\text{occ}$. Inspired by \cite{yuan2018pcn}, we propose a two stage voxel-based PointNet encoder, which can better fuse the global and local geometry inside a voxel. Unlike \cite{yuan2018pcn} who applies maxpooling to all points to get a global embedding, our network computes per-voxel embeddings by maxpooling points inside each voxel. Please see supplementary for details.

\begin{figure}
\begin{center}
   \includegraphics[width=1.0\linewidth]{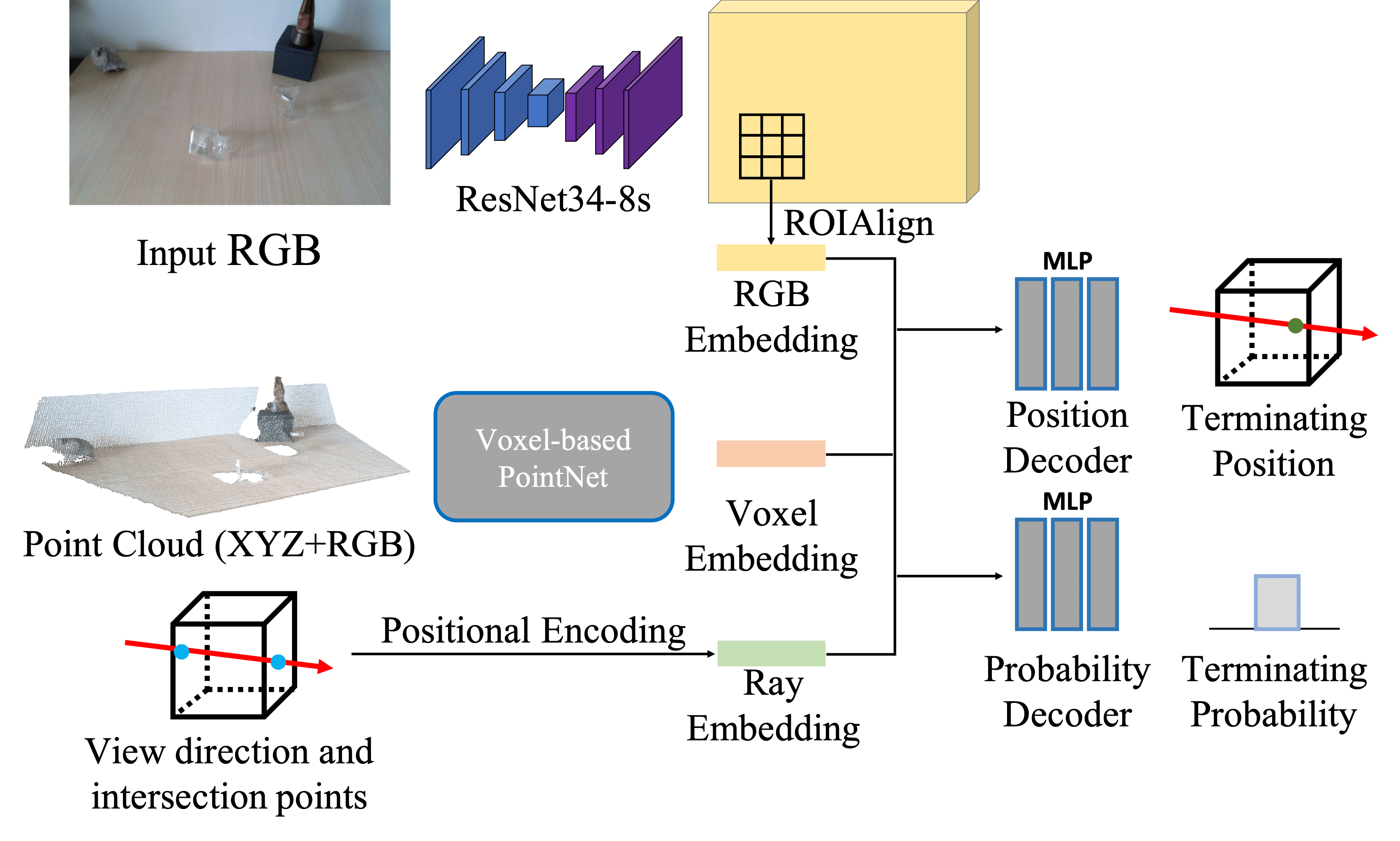}
\end{center}
\vspace{-5mm}
   \caption{LIDF for each ray-voxel pair. Our proposed implicit function queries the termination probability and position given the RGB, voxel and ray embeddings.} 
\label{fig:lidf}
\end{figure}

\subsection{Ray Pooling}
\label{subsec::ray_pool}
Given the predicted terminating probability and position for each ray-voxel pair, we need to estimate the depth of ray $r_i$. We achieve this by applying a pooling operation to all pairs containing $r_i$. There are two natural choices for the pooling: argmax and weighted sum. The argmax operator computes the depth of $r_i$ as the depth estimation of the ray-voxel pair with the maximal ray terminating probability: 
\begin{equation}
    d_i = d_{ij'}, j'=\argmax_{j: v^\text{occ}_{j} \in \mathbf{V}^\text{occ}_{r_i}}p^\text{end}_{ij},
\end{equation}
where $d_i$ is the predicted terminating position of ray $r_i$; $j'$ is the index of the occupied voxel that has the largest terminating probability; $\mathbf{V}^\text{occ}_{r_i}$ represents all the occupied voxels intersected by ray $r_i$.

The weighted sum operator computes the depth of $r_i$ as the sum of depth estimation of all pairs along the ray weighted by their terminating probabilities:
\begin{align}
    d_i = \sum_{j: v^\text{occ}_{j} \in \mathbf{V}^\text{occ}_{r_i}}d_{ij}p^\text{end}_{ij}.
\end{align}
Our experiments, in Section~\ref{subsec::ablative}, show that argmax pooling leads to better accuracy in depth prediction because weighted sum can introduce noise brought by voxels that are far away from the ground truth even if their terminating probabilities are low.

\subsection{Depth Refinement Model}
\label{subsec::drm}
Since the real depth scan around transparent objects can be very noisy or totally missing depth values, the voxel embedding in those areas can encode neither the geometry of transparent objects nor the spatial arrangement between transparent objects and non-transparent ones. To overcome this issue, we propose a self-correcting model to refine the depth estimation progressively. For each iteration, the refinement model recomputes the voxel embedding using input valid points as well as predicted missing points from last iteration, and estimates the correction offset. The forward pass for the $k$-th iteration ($k \geq 1$) can be formulated as:
\begin{align}
& \Hat{d}_{i}[k] = \Hat{d}_{i}[k-1] + \Hat{\delta}_{i}[k] \cdot r_i\\
& \Hat{\delta}_{i}[k] = \Hat{\mathcal{F}}_\text{pos}(\Hat{\mathcal{H}}_{i}[k]) \\
& \Hat{\mathcal{H}}_{i}[k] = \mathcal{H}^\text{rgb}_{i} \oplus \Hat{\mathcal{H}}^\text{vox}_{j_{k-1}}[k] \oplus \gamma(r_i) \oplus \gamma(\Hat{d}_{i}[k-1]),
\end{align}
where $\Hat{d}_{i}[k]$ is the terminating position of ray $r_i$ under $k$-th iteration and we set $\Hat{d}_{i}[0] = d_i$; $\Hat{\delta}_{i}[k]$ is the offset along the ray from $\Hat{d}_{i}[k-1]$ to $\Hat{d}_{i}[k]$; $\Hat{\mathcal{H}}_{i}[k]$ is the input feature embedding for ray $r_i$ under $k$-th iteration; $j_{k-1}$ is the index for the occupied voxel containing position $\Hat{d}_{i}[k-1]$; $\Hat{\mathcal{H}}^\text{vox}_{j_{k-1}}[k]$ is the voxel embedding under $k$-th iteration for voxel $v^\text{occ}_{j_{k-1}}$. We compute $\Hat{\mathcal{H}}^\text{vox}_{j_{k-1}}[k]$ by sending both input valid points and predicted missing points inside $v^\text{occ}_{j_{k-1}}$ into our proposed PointNet encoder. $\Hat{\mathcal{F}}_\text{pos}$ is parametrized as a MLP with iterative error feedback. It is worth noting that all networks in the refinement model do not share weights with the first stage networks. Refinement networks can be applied iteratively to improve the quality of the depth. We apply refinement networks for 2 iterations and refining more than 2 iterations results in marginal improvement of depth quality.

\begin{figure}
\begin{center}
   \includegraphics[width=1.0\linewidth]{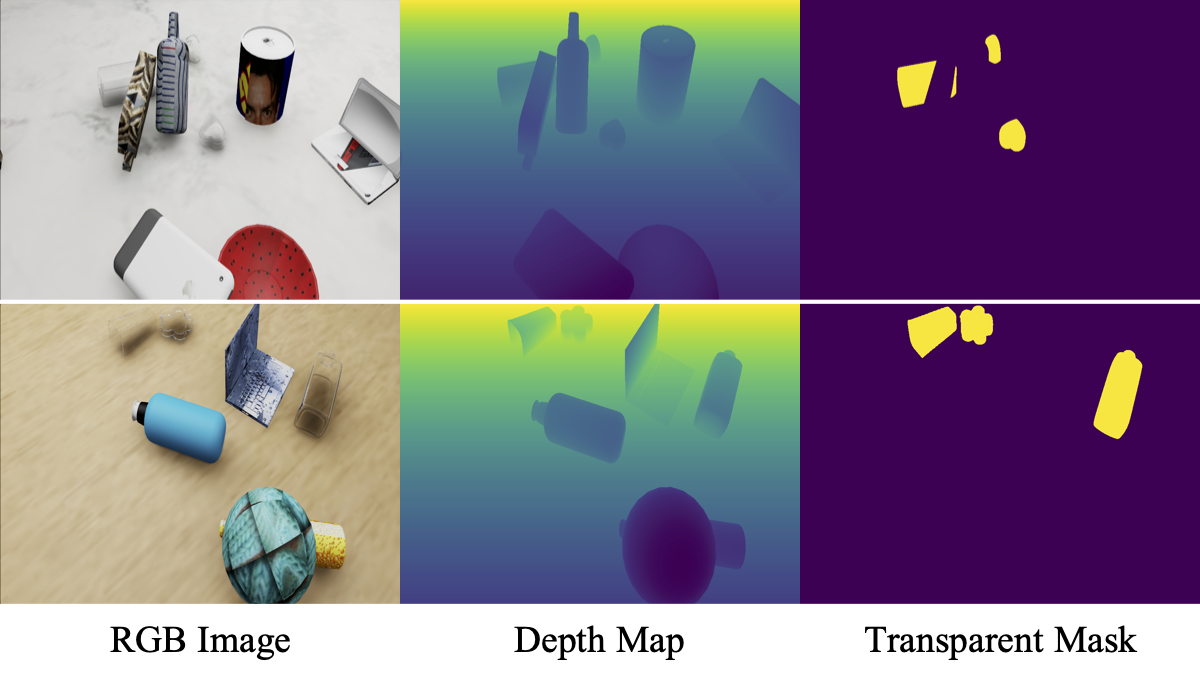}
\end{center}
\vspace{-5mm}
   \caption{Examples from our Omniverse Object Dataset. 
} 
\label{fig:omni_data}
\end{figure}

\subsection{Loss Functions}
Our system is trained using the following loss:
\begin{align}
    \mathcal{L} &= \omega_\text{pos}\mathcal{L}_\text{pos} + \omega_\text{prob}\mathcal{L}_\text{prob} + \omega_\text{sn}\mathcal{L}_\text{sn},
\end{align}
where $\mathcal{L}_\text{pos}$ is the L1 loss between the predicted and groundtruth depth; $\mathcal{L}_\text{prob}$ is the cross entropy loss for terminating probabilities of all intersecting voxels along the ray; $\mathcal{L}_\text{sn}$ is the cosine distance of surface normals computed from predicted and groundtruth point cloud respectively, which can regularize the network to learn meaningful object shapes; $\omega_\text{pos}$, $\omega_\text{prob}$ and $\omega_\text{sn}$ are weights for different losses.

\subsection{Implementation Details}
We set image resolution to $320\times240$ in all experiments. To simulate the noise pattern of real depth scans on our synthetic training data, we remove all depth values for transparent objects, part of the depth values for opaque objects, and create some random holes in the depth map for background. We also augment the color input by adding pixel noise, motion blur and random noise in HSV space. For training, our method only predicts depth for corrupted pixels mentioned above. For testing, depth values of all pixels are estimated so that our method does not rely on the segmentation of transparent objects. Our two stage networks are trained separately. The first stage networks are trained for 60 epochs using Adam optimizer~\cite{kingma2014adam} with a fixed learning rate of 0.001. After that, the first stage networks are frozen and we train the refinement networks for another 60 epochs using Adam optimizer. The first 30 epochs are trained with learning rate 0.001 using the proposed loss function. The later 30 epochs are trained with learning rate 0.0001 using hard negative mining, where only top 10 percent of pixels with largest errors are considered. We set $\omega_\text{pos}=100$ and  $~\omega_\text{sn}=10$ for the first stage and first 30 epochs of the second stage. For the later 30 epochs of second stage, we set $\omega_\text{pos}=20$ and $~\omega_\text{sn}=2$. $~\omega_\text{prob}$ is set to 0.5 for the first stage and 0 for the depth refinement model. 


\section{Experiments}
\label{sec::exp}
\mypar{Datasets} Our full pipeline is trained on the ClearGrasp dataset~\cite{sajjan2020clear} and a new dataset we generated using the Omniverse Platform~\cite{omniverse}, which we call Omniverse Object dataset. The dataset provides various supervisions for transparent and opaque objects in cluterred scenes. Figure~\ref{fig:omni_data} visualizes some examples of the dataset. 3D object models in Omniverse Object dataset are collected from ClearGrasp and ShapeNet~\cite{shapenet2015}.  To get natural poses of objects, we use NVIDIA PhysX engine to simulate objects falling to the ground. Then we randomly select some objects and set their materials to the glass. We also augment the data by changing textures for the ground and opaque objects, lighting conditions and camera views. See supplementary for more details about Omniverse Object dataset. The evaluation is done on the ClearGrasp dataset~\cite{sajjan2020clear}. It has 4 types of different testing data: Synthetic images of 5 training objects (Syn-known); Synthetic images of 4 novel objects (Syn-novel); Real world images of 5 training objects (Real-known); Real world images of 5 novel objects (Real-novel), 3 of them are not present in synthetic data. 

\mypar{Metrics} We closely follow the evaluation protocal of ~\cite{sajjan2020clear} for the depth estimation. The prediction and groundtruth are first resized to $144 \times 256$ resolution, then we compute errors in the transparent objects area using the following metrics:
{
\small
\begin{tabular}{l}
Root Mean Squared Error (RMSE): $\sqrt{\frac1{|\Hat{\mathcal{D}}|}\sum_{d\in \Hat{\mathcal{D}}}||d - d^*||^2}$ \\
Absolute Relative Difference (REL):  $\frac1{|\Hat{\mathcal{D}}|}\sum_{d\in \Hat{\mathcal{D}}}|d - d^*| / d^*$ \\
Mean Absolute Error (MAE): $\frac1{|\Hat{\mathcal{D}}|}\sum_{d\in \Hat{\mathcal{D}}}|d - d^*|$ \\
Threshold: \% of $d_{i}$ satisfying $\max(\frac{d_i}{d_i^*},\frac{d_i^*}{d_i}) < \delta $ \\
\end{tabular}
}
The first 3 metrics are computed in meters. For the threshold, $\delta$ is set to 1.05, 1.10 and 1.25.

\begin{figure*}
\begin{center}
   \includegraphics[width=0.95\linewidth]{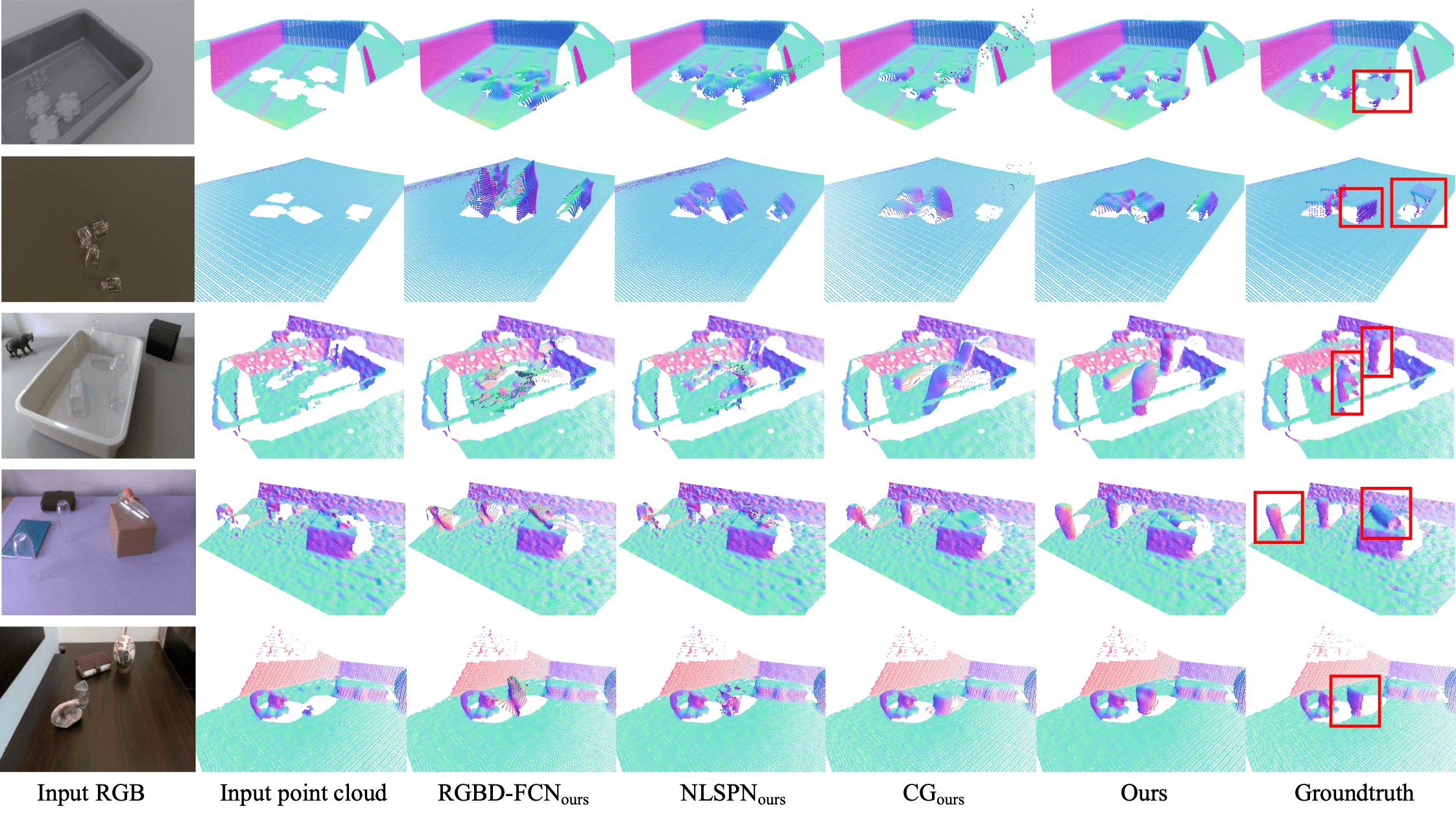}
\end{center}
\vspace{-5mm}
   \caption{Qualitative comparison to state-of-the-art methods. For all the baselines, we provide visualizations of the version retrained on our data for fair comparison. The point clouds are colored by surface normal and rendered in a novel viewpoint to better visualize the 3D shape. Red boxes in the groundtruth highlight areas where our method performs much better than all baselines. Zoom in to see details.} 
\label{fig:qual_sota}
\end{figure*}

\subsection{Comparison to State-of-the-art Methods}
We compare our approach to several state-of-the-art methods in Table~\ref{table:sota}. For fair comparison, we evaluate all related works using their released checkpoints (denoted by method name) as well as retraining on our data (denoted by method name with subscript ours. 
All baselines are trained on both datasets together which is the same setting as our proposed method). 
RGBD-FCN$_\text{ours}$ is a strong baseline proposed by ourselves. It directly regresses depth maps using fully convolutional networks from RGB-D images. We use Resnet34-8s~\cite{xiang2020learning} as the network architecture and train the network on our data. 
NLSPN~\cite{park2020non} is the state-of-the-art method for depth completion on NYUV2~\cite{Silberman:ECCV12} and KITTI~\cite{Uhrig2017THREEDV} dataset. 
Cleargrasp ~\cite{sajjan2020clear} is the state-of-the-art method for depth completion of transparent objects. 
For our approach, we use the best model: LIDF plus the depth refinement model. 
Our method achieves the best result on all datasets even when baseline methods are trained on the same data. It also shows that training on Omniverse Object dataset can boost the performance of baseline methods. 
In Figure~\ref{fig:qual_sota}, we provide qualitative comparison by rendering point cloud in a novel view. Our approach can generate more meaningful depth than baseline methods. 

Inference speed is an important factor for deploying the model in real applications. We compare the inference time of our method to the best performing baseline ClearGrasp~\cite{sajjan2020clear}. While our method provides an end-to-end solution to complete the depth map of the scene, ClearGrasp uses a two-stage approach where the first stage uses CNN to predict some intermediate observations and the second stage solves an expensive optimization to recover the missing depths. In order to compare the runtime fairly, we run both methods on a machine equipped with a single P100 GPU and compute the average time over the whole testing set. Our method takes \textbf{0.09} second for one image while ClearGrasp needs to spend \textbf{1.82} second, showing that our method is \textbf{20x} faster.

\begin{table}[ht]
\setlength\tabcolsep{1.5pt}
\centering
\scalebox{0.95}{
\begin{tabular}{l|cccccc}
\toprule
\multicolumn{1}{c|}{Methods} & RMSE$\downarrow$ & REL$\downarrow$ & MAE$\downarrow$  & $\delta_{1.05}\uparrow$ & $\delta_{1.10}\uparrow$ & $\delta_{1.25}\uparrow$ \\
\hline
\multicolumn{1}{c|}{} & \multicolumn{6}{c}{Cleargrasp Syn-known} \\
\hline
RGBD-FCN$_\text{ours}$ & 0.028 & 0.039 & 0.021 & 76.53 & 91.82 & 99.00 \\
NLSPN~\cite{park2020non} & 0.136 & 0.231 & 0.113 & 19.02 & 35.95 & 70.43 \\
NLSPN$_\text{ours}$~\cite{park2020non} & 0.026 & 0.041 & 0.021 & 74.89 & 89.95 & 98.59 \\
CG~\cite{sajjan2020clear} & 0.041 & 0.055 & 0.031 & 69.43 & 89.17 & 96.74 \\
CG$_\text{ours}$~\cite{sajjan2020clear} & 0.034 & 0.045 & 0.026 & 73.53 & 92.68 & 98.25 \\
Ours          & \textbf{0.012} & \textbf{0.017} & \textbf{0.009} & \textbf{94.79} & \textbf{98.52} & \textbf{99.67} \\

\hline
\multicolumn{1}{c|}{} & \multicolumn{6}{c}{Cleargrasp Syn-novel} \\
\hline
RGBD-FCN$_\text{ours}$ & 0.033 & 0.058 & 0.028 & 52.40 & 85.64 & 98.94 \\
NLSPN~\cite{park2020non} & 0.132 & 0.239 & 0.106 & 16.25 & 32.13 & 64.78 \\
NLSPN$_\text{ours}$~\cite{park2020non} & 0.029 & 0.049 & 0.024 & 64.83 & 88.20 & 98.57 \\
CG~\cite{sajjan2020clear} & 0.044 & 0.074 & 0.038 & 41.37 & 79.20 & 97.29 \\
CG$_\text{ours}$~\cite{sajjan2020clear} & 0.037 & 0.062 & 0.032 & 50.27 & 84.00 & 98.39 \\
Ours          & \textbf{0.028} & \textbf{0.045} & \textbf{0.023} & \textbf{68.62} & \textbf{89.10} & \textbf{99.20} \\

\hline
\multicolumn{1}{c|}{} & \multicolumn{6}{c}{Cleargrasp Real-known} \\
\hline
RGBD-FCN$_\text{ours}$ & 0.054 & 0.087 & 0.048 & 36.32 & 67.11 & 96.26 \\
NLSPN~\cite{park2020non} & 0.149 & 0.228 & 0.127 & 14.04 & 26.67 & 54.32 \\
NLSPN$_\text{ours}$~\cite{park2020non} & 0.056 & 0.086 & 0.048 & 40.60 & 67.68 & 96.25 \\
CG~\cite{sajjan2020clear} & 0.039 & 0.051 & 0.029 & 72.62 & 86.96 & 95.58 \\
CG$_\text{ours}$~\cite{sajjan2020clear} & 0.032 & 0.042 & 0.024 & 74.63 & 90.69 & 98.33 \\
Ours          & \textbf{0.028} & \textbf{0.033} & \textbf{0.020} & \textbf{82.37} & \textbf{92.98} & \textbf{98.63} \\

\hline
\multicolumn{1}{c|}{} & \multicolumn{6}{c}{Cleargrasp Real-novel} \\
\hline
RGBD-FCN$_\text{ours}$ & 0.042 & 0.070 & 0.037 & 42.45 & 75.68 & 99.02 \\
NLSPN~\cite{park2020non} & 0.145 & 0.240 & 0.123 & 13.77 & 25.81 & 51.59 \\
NLSPN$_\text{ours}$~\cite{park2020non} & 0.036 & 0.059 & 0.030 & 51.97 & 84.82 & 99.52 \\
CG~\cite{sajjan2020clear} & 0.034 & 0.045 & 0.025 & 76.72 & 91.00 & 97.63 \\
CG$_\text{ours}$~\cite{sajjan2020clear} & 0.027 & 0.039 & 0.022 & \textbf{79.5} & 93.00 & 99.28 \\
Ours          & \textbf{0.025} & \textbf{0.036} & \textbf{0.020} & 76.21 & \textbf{94.01} & \textbf{99.35} \\
\bottomrule
\end{tabular}}
\caption{Quantitative Comparison to state-of-the-art methods. $\downarrow$ means lower is better, $\uparrow$ means higher is better. Please refer to the text for more details.}
\label{table:sota}
\end{table}


\subsection{Ablation Studies}
\label{subsec::ablative}
In this section, we first evaluate the effect of the depth refinement model. After that, we compare several configurations for our first stage networks. To focus on the generalization ability, we only report quantitative results on ClearGrasp Real-novel dataset. Please refer to the supplementary for results on other testing data.

\begin{figure*}
\begin{center}
   \includegraphics[width=0.95\linewidth]{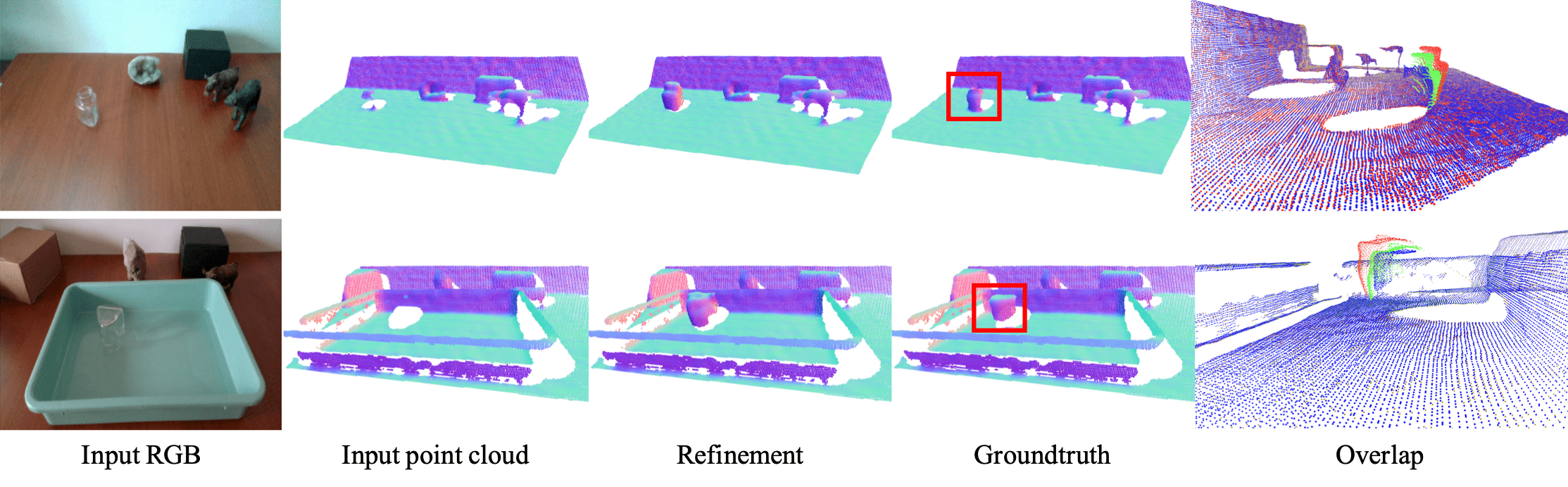}
\end{center}
\vspace{-5mm}
   \caption{Examples of the depth refinement model. The point clouds (column 2-4) are colored by surface normal and rendered in a novel viewpoint to better visualize the 3D shape. The red box in the Groundtruth highlights the interest area. The last column visualizes the overlap of point clouds for non-refinement (red), refinement (green) and groundtruth (blue). Please zoom in to see details.} 
\label{fig:qual_refine}
\vspace{-3mm}
\end{figure*}

\mypar{Depth refinement model.} To demonstrate the effectiveness of the depth refinement model, we run two experiments with and without refinement. Table~\ref{table:refine} shows that adding depth refinement can boost the performance on real unseen objects. In Figure~\ref{fig:qual_refine}, we can see that refinement model can correct the tilted error from the non-refinement prediction.

\begin{table}[ht]
\setlength\tabcolsep{1.5pt}
\centering
\begin{tabular}{c|cccccc}
\toprule
Refinement & RMSE$\downarrow$ & REL$\downarrow$ & MAE$\downarrow$  & $\delta_{1.05}\uparrow$ & $\delta_{1.10}\uparrow$ & $\delta_{1.25}\uparrow$ \\
\hline
$\times$ & 0.028 & 0.043 & 0.023 & 65.17 & 92.23 & 99.30 \\
\checkmark & \textbf{0.025} & \textbf{0.036} & \textbf{0.020} & \textbf{76.21} & \textbf{94.01} & \textbf{99.35} \\
\bottomrule
\end{tabular}
    \caption{Depth refinement model. $\times$ denotes prediction from first stage while \checkmark denotes prediction from refinement model.}
\label{table:refine}
\end{table}

\begin{table}[ht]
\setlength\tabcolsep{1.5pt}
\centering
\vspace{-2mm}
\begin{tabular}{cc|cccccc}
\toprule
RGB & Depth & RMSE$\downarrow$ & REL$\downarrow$ & MAE$\downarrow$  & $\delta_{1.05}\uparrow$ & $\delta_{1.10}\uparrow$ & $\delta_{1.25}\uparrow$ \\
\hline
\checkmark & \checkmark &  \textbf{0.028} & \textbf{0.043} & \textbf{0.023} & \textbf{65.17} & \textbf{92.23} & \textbf{99.30} \\
& \checkmark & 0.052 & 0.084 & 0.043 & 39.14 & 66.49 & 97.03 \\
\checkmark & &  0.062 & 0.096 & 0.054 & 35.02 & 55.98 & 93.69 \\
\bottomrule
\end{tabular}
    \caption{Ablation studies for effect of different modalities}
\label{table:embeddings}
\vspace{-5mm}
\end{table}

\mypar{Input Modalities.} Our method uses combination of RGB and depth information to complete the depth map. To evaluate the contribution of each input modality, we retrained the first stage model using only one of modalities. Table~\ref{table:embeddings} shows how much each modality contributes to the accuracy of the predicted depth. RGB alone becomes similar to depth estimation from single image and it cannot predict the depth accurately. In addition, depth alone is not able to predict depth for transparent objects well because transparent objects are not observed in depth and without RGB information the model is agnostic of transparent objects.

\begin{table}[ht]
\setlength\tabcolsep{1.5pt}
\centering
\vspace{-2mm}
\scalebox{0.8}{
\begin{tabular}{ccc|cccccc}
\toprule
Ray Info & Pos. Enc & \# Vox & RMSE$\downarrow$ & REL$\downarrow$ & MAE$\downarrow$  & $\delta_{1.05}\uparrow$ & $\delta_{1.10}\uparrow$ & $\delta_{1.25}\uparrow$ \\
\hline
\checkmark & \checkmark & $8^3$ & \textbf{0.028} & \textbf{0.043} & \textbf{0.023} & \textbf{65.17} & \textbf{92.23} & 99.30 \\
 & N/A & $8^3$ & 0.050 & 0.064 & 0.035 & 57.88 & 80.29 & 95.87 \\
\checkmark & & $8^3$ & 0.032 & 0.050 & 0.026 & 60.96 & 83.86 & 99.21 \\
\checkmark & \checkmark & $4^3$ & 0.030 & 0.046 & 0.025 & 64.00 & 88.96 & \textbf{99.66} \\
\checkmark & \checkmark & $16^3$ & 0.038 & 0.058 & 0.031 & 56.78 & 81.49 & 96.21 \\
\bottomrule
\end{tabular}}
    \caption{Ablation Study for different design choices such as including ray information in the embedding, applying positional encoding, and the number of voxels on the accuracy}
\label{table:design_choices} \vspace{-4mm}
\end{table}

\mypar{Ray Information.} To further highlight the effectiveness of our ray-voxel formulation, we compared our method with the variation where RGB embedding and voxel embedding are concatenated with each other and there is no information about the ray (ray-voxel intersection points and ray direction) provided to the model. Table~\ref{table:design_choices} shows that ray information provides crucial guidance to the model to predict the depth effectively.

\mypar{Positional encoding.} Positional encoding was shown to be quite important in~\cite{mildenhall2020nerf} to capture high frequency details. We also conducted an experiment by disabling the positional encoding. Table~\ref{table:design_choices} shows that positional encoding is making significant changes in $\delta_{1.05}$ and $\delta_{1.10}$ which reflect the ability of the model to capture fine details.

\mypar{Number of Voxels.} We explore the sensitivity of our method to the number of voxels. Smaller number of voxels makes ray termination classification easier but regression for the offset more challenging while larger number of voxels leads to the opposite. Table~\ref{table:design_choices} shows that if we make the number of voxels too large the performance drops more significantly compared to the smaller number of voxels.

\begin{table}[ht]
\setlength\tabcolsep{1.5pt}
\centering

\begin{tabular}{c|cccccc}
\toprule
Data & RMSE$\downarrow$ & REL$\downarrow$ & MAE$\downarrow$  & $\delta_{1.05}\uparrow$ & $\delta_{1.10}\uparrow$ & $\delta_{1.25}\uparrow$ \\
\hline
CG+Omni & \textbf{0.028} & \textbf{0.043} & \textbf{0.023} & \textbf{65.17} & \textbf{92.23} & 99.30 \\
Omni & 0.037 & 0.057 & 0.030 & 52.82 & 85.16 & \textbf{99.43} \\
CG & 0.038 & 0.057 & 0.031 & 58.95 & 80.13 & 96.87 \\
\bottomrule
\end{tabular}
    \caption{Quantitative effect of training data on the generalization to real novel objects.}
\vspace{-4mm}
\label{table:training_data}
\end{table}

\mypar{Training Data.} We analyze the effects of training data in Table~\ref{table:training_data}. We find training our method purely on ClearGrasp or Omniverse leads to similar results, but training on both datasets can improve the performance a lot. This indicates that Omniverse dataset can be a good complementary to ClearGrasp for transparent objects learning.

\begin{figure}
\begin{center}
   \includegraphics[width=1.0\linewidth]{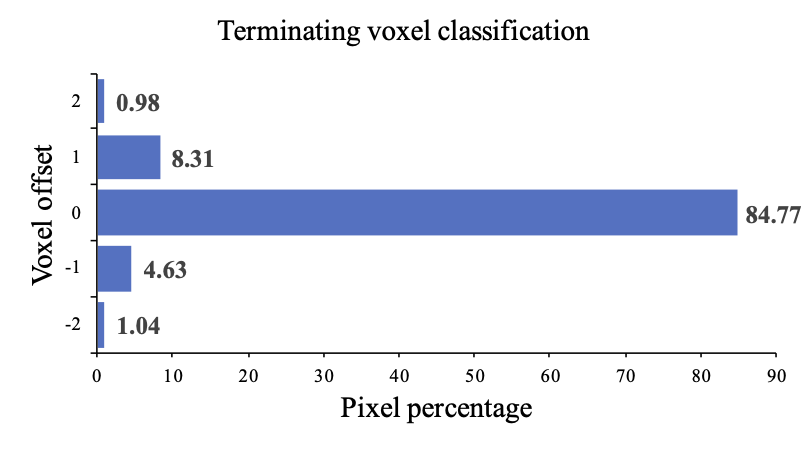}
\end{center}
\vspace{-5mm}
   \caption{Distribution of the terminating voxel classification. Vertical axis represents the offset from the groundtruth to the predicted terminating voxel along the ray. For example, +1(-1) represents the predicted terminating voxel is 1 voxel behind (before) the groundtruth along the ray. Horizontal axis represents the percentage of pixels falls into that category.}
\label{fig:hist_classify}
\end{figure}

\mypar{Ray Pooling.} Table~\ref{table:ray_pooling} provides the comparison for two types of pooling operators described in section~\ref{subsec::ray_pool}. The argmax operator performs better than weighted sum operator on real-novel objects. We argue it is mainly because weighted sum operator suffers from the noise introduced by voxels far away from the ground truth. Figure~\ref{fig:hist_classify} provides the accuracy histogram of the terminating voxel classification for the argmax operator. We can see that 84.77\% of pixels are correctly classified and 97.71\% of pixels are classified within $\pm1$ voxel. We further compute the distance from the ground truth to the misclassfied $\pm1$ voxels. The mean distance from the ground truth position to the leaving (entering) voxel position along the ray for $-1$ ($+1$) voxel is 0.064 (0.048) meter, indicating that groundtruth is very close to the misclassified $-1$ ($+1$) voxel. This implies that erroneous ending voxel classification will not hurt the performance much for the argmax operator. 

\begin{table}[ht]
\setlength\tabcolsep{1.5pt}
\centering
\vspace{-2mm}
\scalebox{0.95}{
\begin{tabular}{c|cccccc}
\toprule
Ray Pooling & RMSE$\downarrow$ & REL$\downarrow$ & MAE$\downarrow$  & $\delta_{1.05}\uparrow$ & $\delta_{1.10}\uparrow$ & $\delta_{1.25}\uparrow$ \\
\hline
Argmax & \textbf{0.028} & \textbf{0.043} & \textbf{0.023} & \textbf{65.17} & \textbf{92.23} & \textbf{99.30} \\
WeightedSum & 0.032 & 0.051 & 0.027 & 59.31 & 84.25 & 98.34 \\
\bottomrule
\end{tabular}}
    \caption{Ablation study for different ray pooling strategies.}
\label{table:ray_pooling}
\vspace{-2mm}
\end{table}

\mypar{Candidate points selection.} To get the depth of the camera ray, our method learns the offset inside each voxel and apply argmax ray pooling while NeRF~\cite{mildenhall2020nerf} tries to densely sample points and apply volume rendering equation. We slightly modify NeRF by sampling points only inside intersecting voxels. Table~\ref{table:cand_points} shows that learning the position of candidate points is better than heuristic sampling strategy.

\begin{table}[ht]
\setlength\tabcolsep{1.5pt}
\centering
\scalebox{0.95}{
\begin{tabular}{c|cccccc}
\toprule
Candidates & RMSE$\downarrow$ & REL$\downarrow$ & MAE$\downarrow$  & $\delta_{1.05}\uparrow$ & $\delta_{1.10}\uparrow$ & $\delta_{1.25}\uparrow$ \\
\hline
Learned offset & \textbf{0.028} & \textbf{0.043} & \textbf{0.023} & \textbf{65.17} & \textbf{92.23} & \textbf{99.30} \\
Sample points & 0.038 & 0.058 & 0.031 & 54.09 & 79.48 & 98.28 \\
\bottomrule
\end{tabular}}
    \caption{Ablation study for candidate points selection.}
\label{table:cand_points}
\vspace{-5mm}
\end{table}


\section{Conclusion}
\label{sec::conclusion}
We have presented a novel framework for depth completion of transparent objects. Our method consists of a local implicit depth function defined on ray-voxel pairs and an iterative depth refinement model. We also introduce a large scale synthetic dataset for transparent objects learning, which can boost the performance for both our approach and other competing methods. Our pipeline is only trained on synthetic datasets but can generalize well to real world scenarios. We thoroughly evaluated our method compared to prior art and ablation baselines. Both quantitative and qualitative results demonstrate substantial improvements over the state-of-the-art in terms of accuracy and speed.

{\small
\bibliographystyle{ieee_fullname}
\bibliography{egbib}
}

\newpage
\clearpage
\appendix

\makeatletter

\section*{Appendix}

\section{Voxel-based PointNet Encoder}
\begin{figure}
\begin{center}
   \includegraphics[width=1.0\linewidth]{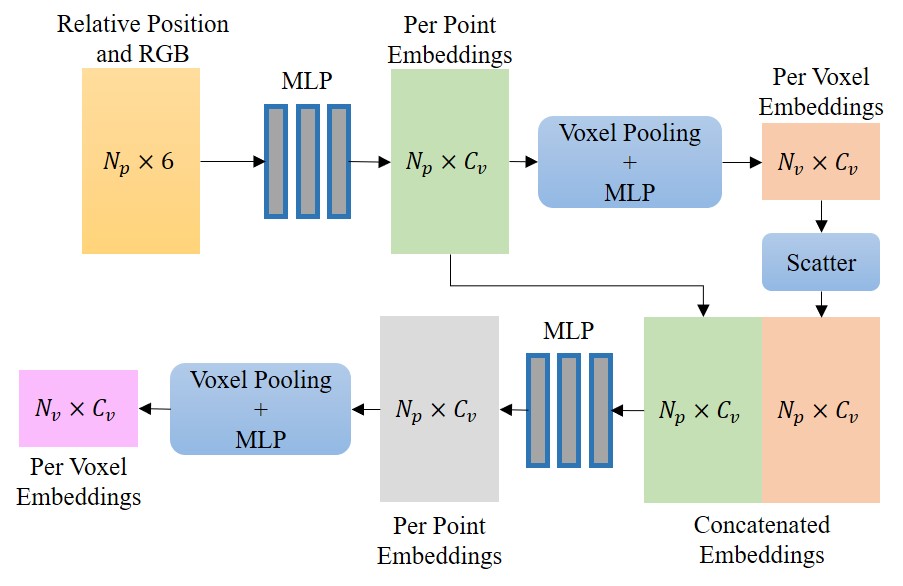}
\end{center}
   \caption{voxel-based PointNet Encoder.} 
\label{fig:vox_embed}
\end{figure}

In this section, we provide more details about proposed two-stage voxel-based PointNet encoder. As illustrated in Figure~\ref{fig:vox_embed}, We first compute the relative position of valid points $P^\text{valid}$ with respect to the center of voxels they reside in. The relative position and color of valid points are sent into a shared MLP to produce the initial per-point embedding $\Tilde{\mathcal{H}}^\text{pcl} \in \mathbb{R}^{N_p \times C_v}$, where $N_p$ is the number of valid points and $C_v$ is the dimension of voxel embedding. Then we apply the max-pooling to embeddings of all points inside each voxel, followed by another MLP to get initial per-voxel embedding $\Tilde{\mathcal{H}}^\text{vox} \in \mathbb{R}^{N_v \times C_v}$, where $N_v$ is the number of occupied voxels. To generate the second stage input for the i-th valid point $P^\text{valid}_i$, we concatenate the point embedding $\Tilde{\mathcal{H}}^\text{pcl}_i$ and the voxel embedding $\Tilde{\mathcal{H}}^\text{vox}_k$, satisfying $P^\text{valid}_i$ reside in $v^\text{occ}_{k}$. Finally, we feed in the new input and repeat the same process as the first stage to get the voxel embedding $\mathcal{H}^\text{vox} \in \mathbb{R}^{N_v \times C_v}$.

\section{Omniverse Object Dataset}
In this section, we provide more details about our Omniverse Object Dataset. To generate the dataset, following categories from ShapeNet~\cite{shapenet2015} are chosen: phone, bowl, camera, laptop, can, bottle. Following objects from ClearGrasp dataset~\cite{sajjan2020clear} are chosen: cup-with-waves, flower-bath-bomb, heart-bath-bomb, square-plastic-bottle, stemless-plastic-champagne-glass. Note that we only select training objects from ClearGrasp dataset to make sure testing objects are never seen during training.
The background textures are randomly selected from the CC0 TEXTURES Dataset~\cite{cc0textures}. The textures for opaque objects are randomly selected from CC0 TEXTURES Dataset~\cite{cc0textures} and Describable Textures Dataset~\cite{cimpoi14describing}.

For each image, we provide the following groundtruth data: depth map, instance segmentation, transparent object segmentation, intrinsic and extrinsic camera parameters, 2d/3d bounding box for each object, 6D pose for each object.
Since the depth map created from ray-tracing is not accurate for transparent objects, we utilize a two-pass rendering strategy to solve it. Before the rendering, we randomly select some objects and list them as transparent candidates. During the first pass, materials of all objects are set to opaque and we render all groundtruth data including depth map using real time ray-tracing. During the second pass, we set materials of transparent candidates to glass and render the RGB image using path tracing.

\section{Additional Results for Ablation Studies}
In this section, we provide quantitative results of ablation studies on ClearGrasp~\cite{sajjan2020clear} Syn-known, Syn-novel and Real-Known dataset. We also provide qualitative comparison of ablation studies on real images.

\mypar{Depth refinement model.} Table~\ref{table:refine_supp} shows that depth refinement model can boost the performance of synthetic novel objects while achieving similar results on synthetic known and real known objects. This further proves that depth refinement model can increase the generality of our approach.

\begin{figure*}
\begin{center}
   \includegraphics[width=1.0\linewidth]{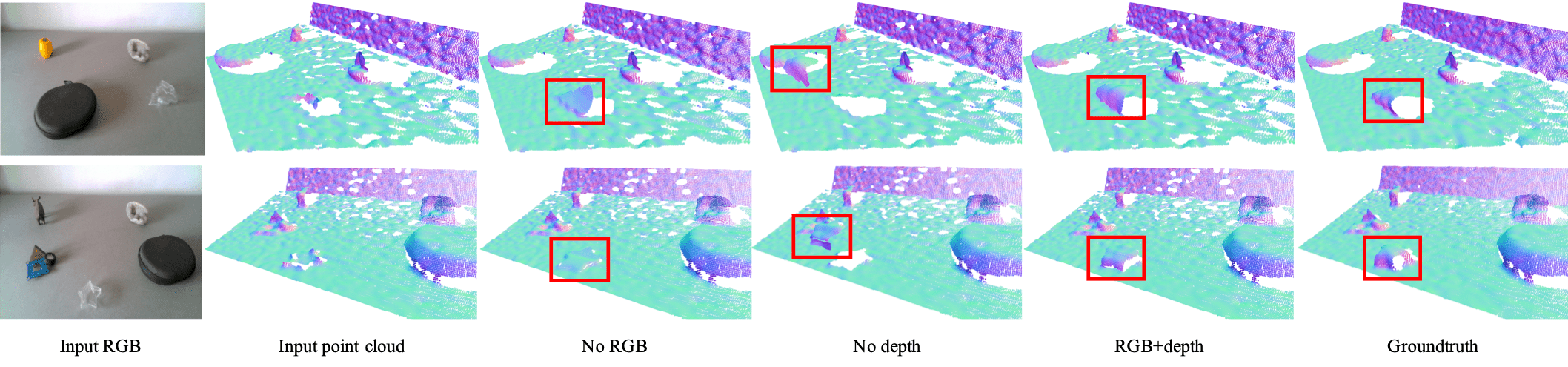}
\end{center}
\vspace{-3mm}
   \caption{Qualitative results for input modalities. Point clouds are colored by surface normal and rendered in a novel viewpoint to better visualize the 3D shape. The red boxes highlights the interest area. Please zoom in to see details.} 
\label{fig:input_modality}
\end{figure*}

\mypar{Input Modalities.} Table~\ref{table:embeddings_supp} shows that both RGB and depth information contribute a lot to the depth accuracy. In Figure~\ref{fig:input_modality}, we also provide qualitative comparison of input modalities on real images. RGB information can provide visual cues about object shapes. Our approach can only predict flat planes without RGB input. The depth information can help localize the object in metric space. The prediction is far from the table without depth input.

\begin{figure*}
\begin{center}
   \includegraphics[width=1.0\linewidth]{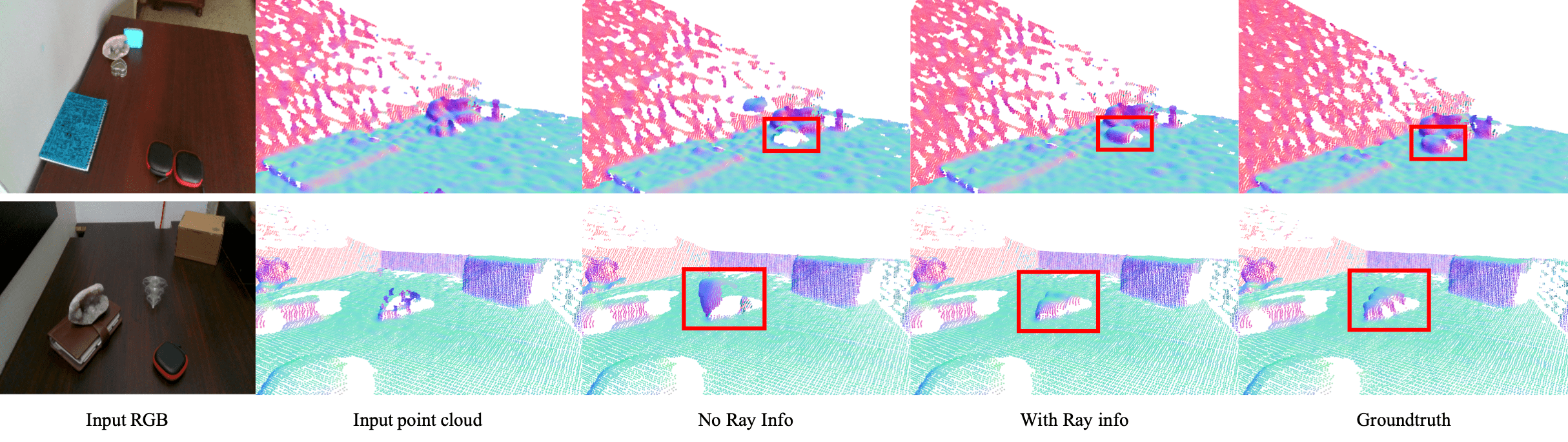}
\end{center}
\vspace{-3mm}
   \caption{Qualitative results for ray information. Point clouds are colored by surface normal and rendered in a novel viewpoint to better visualize the 3D shape. The red boxes highlights the interest area. Please zoom in to see details.} 
\label{fig:ray_info}
\end{figure*}

\mypar{Ray Information.} Table~\ref{table:design_choices_supp} (row 1 and 2 in every sub-table) provides quantitative results of the ray information. Figure~\ref{fig:ray_info} further visualizes some examples with and without ray information as input. We can see that ray information can help the model reason about the location and orientation of transparent objects.

\begin{figure*}
\begin{center}
   \includegraphics[width=1.0\linewidth]{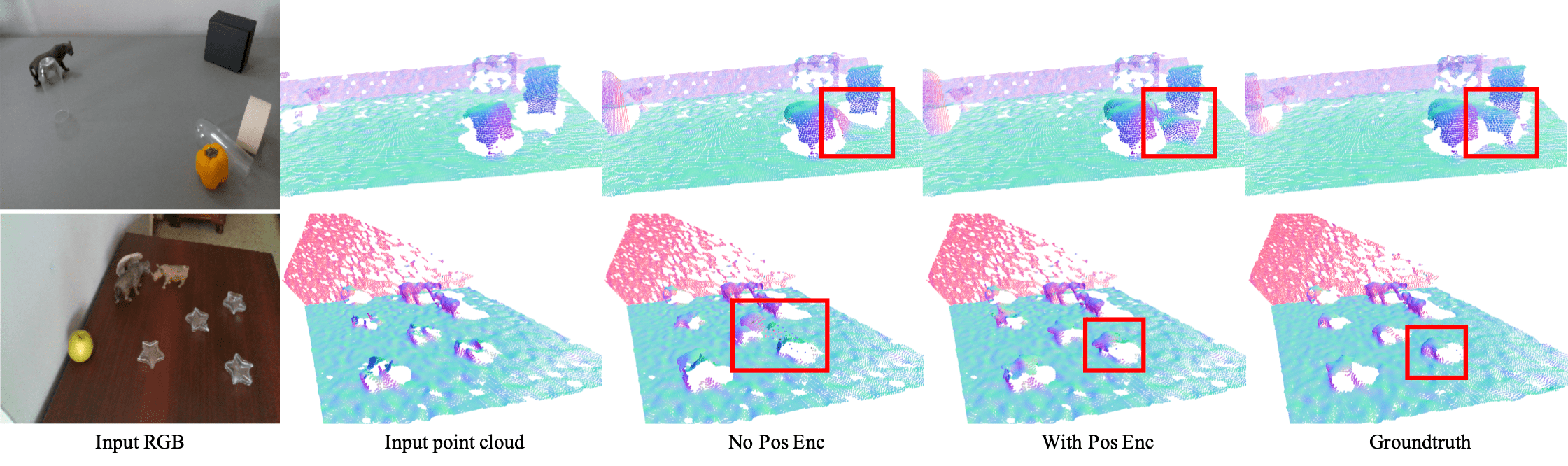}
\end{center}
\vspace{-3mm}
   \caption{Qualitative results for positional encoding. Point clouds are colored by surface normal and rendered in a novel viewpoint to better visualize the 3D shape. The red boxes highlights the interest area. Please zoom in to see details.} 
\label{fig:pos_enc}
\end{figure*}

\mypar{Positional encoding.} Table~\ref{table:design_choices_supp} (row 1 and 3 in every sub-table) shows that positional encoding can improve the performance on both synthetic and real cases. Figure~\ref{fig:pos_enc} shows that positional encoding helps the model to learn fine details of small objects or under heavy occlusion.

\begin{figure*}
\begin{center}
   \includegraphics[width=1.0\linewidth]{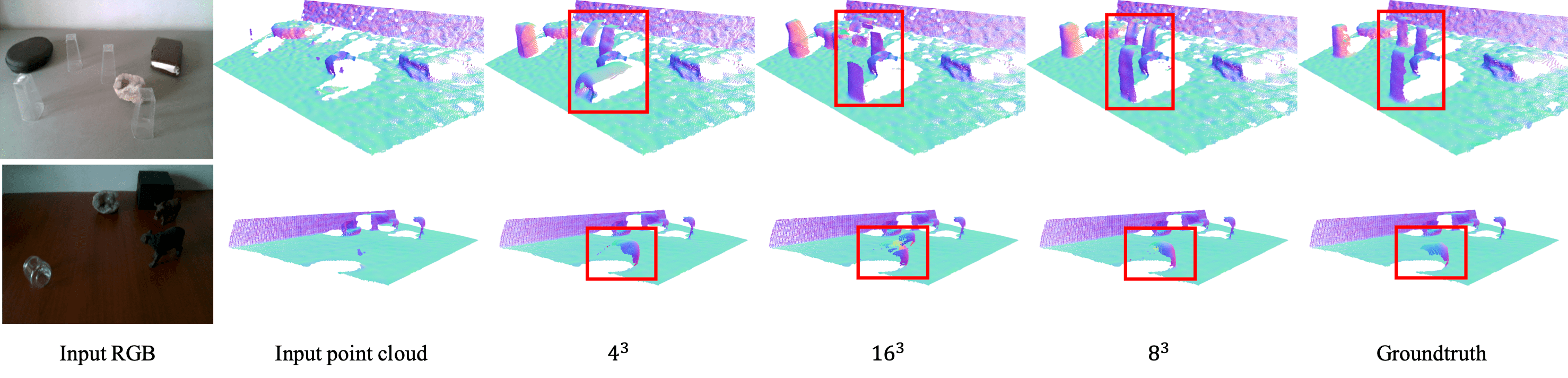}
\end{center}
\vspace{-3mm}
   \caption{Qualitative results for number of voxels. Point clouds are colored by surface normal and rendered in a novel viewpoint to better visualize the 3D shape. The red boxes highlights the interest area. Please zoom in to see details.} 
\label{fig:grid_size}
\end{figure*}

\mypar{Number of voxels.} Table~\ref{table:design_choices_supp} (row 1,4,5 in every sub-table) shows that the accuracy will drop a lot if the number of voxels is too large. Figure~\ref{fig:grid_size} provides predictions of real images under various number of voxels. Smaller number of voxels leads to harder offset regression and the orientation of objects might be wrong (first row). Larger number of voxels causes objects splitting because of harder classification (second row).

\begin{figure*}
\vspace{-5mm}
\begin{center}
   \includegraphics[width=1.0\linewidth]{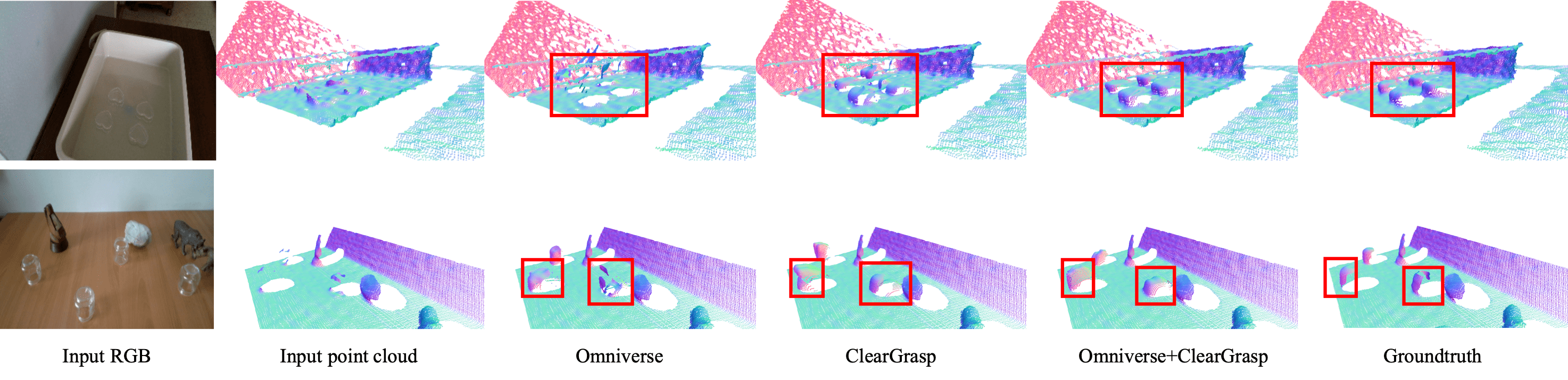}
\end{center}
\vspace{-3mm}
   \caption{Qualitative results for training data. Point clouds are colored by surface normal and rendered in a novel viewpoint to better visualize the 3D shape. The red boxes highlights the interest area. Please zoom in to see details.} 
\label{fig:training_data}
\end{figure*}

\mypar{Training Data.} Table~\ref{table:training_data_supp} and Figure~\ref{fig:training_data} provide quantitative and qualitative comparison on different training data respectively. We can see that training the model on both datasets can get best results.

\begin{figure*}
\begin{center}
   \includegraphics[width=1.0\linewidth]{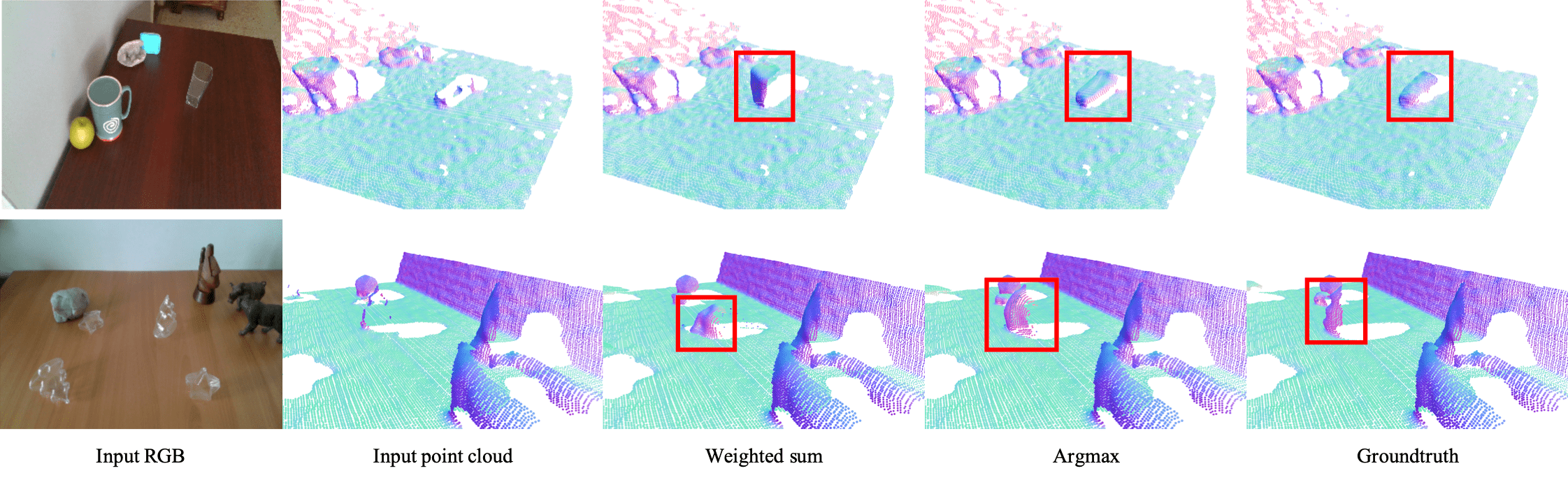}
\end{center}
\vspace{-3mm}
   \caption{Qualitative results for ray pooling. Point clouds are colored by surface normal and rendered in a novel viewpoint to better visualize the 3D shape. The red boxes highlights the interest area. Please zoom in to see details.} 
\label{fig:ray_pooling}
\end{figure*}

\mypar{Ray Pooling.} Table~\ref{table:ray_pooling_supp} shows that argmax performs consistently better than weighted sum on all types of testing data. Figure~\ref{fig:ray_pooling} also shows that argmax can better estimate missing depth of transparent objects on real images.

\begin{figure*}
\begin{center}
   \includegraphics[width=1.0\linewidth]{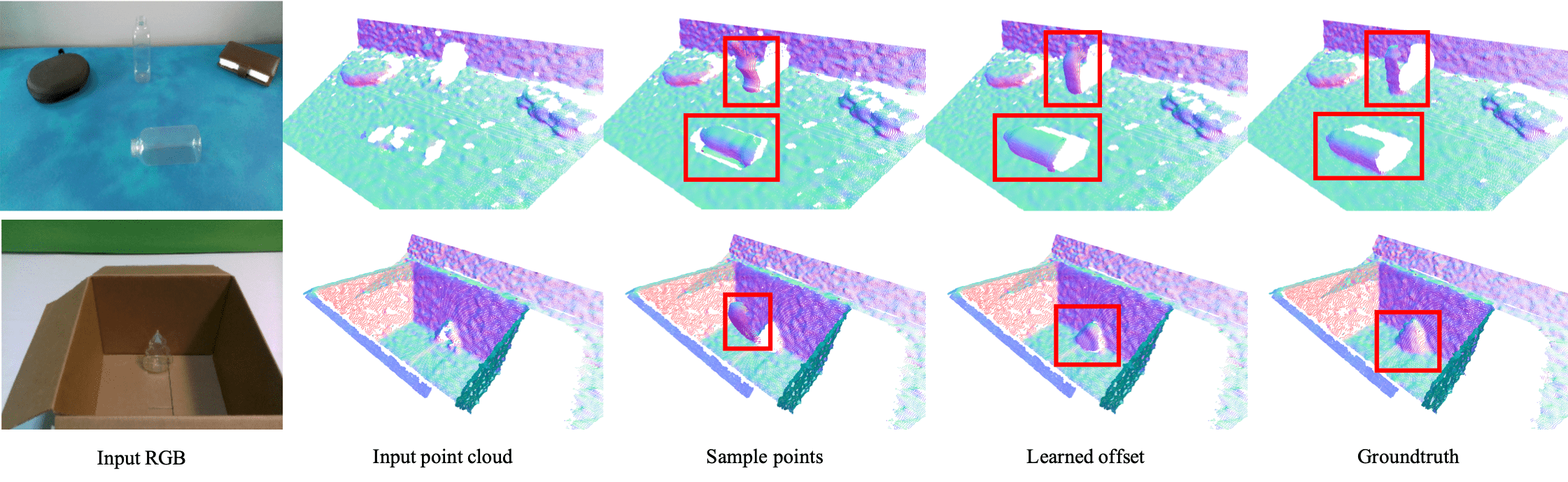}
\end{center}
\vspace{-3mm}
   \caption{Qualitative results for candidate points selection. Point clouds are colored by surface normal and rendered in a novel viewpoint to better visualize the 3D shape. The red boxes highlights the interest area. Please zoom in to see details.} 
\label{fig:cand_points}
\end{figure*}

\mypar{Candidate points selection.} Table~\ref{table:cand_points_supp} shows that directly learning offsets of candidate points is better than sampling points heuristically. Figure~\ref{fig:cand_points} further provides some examples on real images, showing that learning offset is more robust to strong background textures.

\section{Qualitative Results on NYUV2 Dataset}
\begin{figure*}
\begin{center}
   \includegraphics[width=1.0\linewidth]{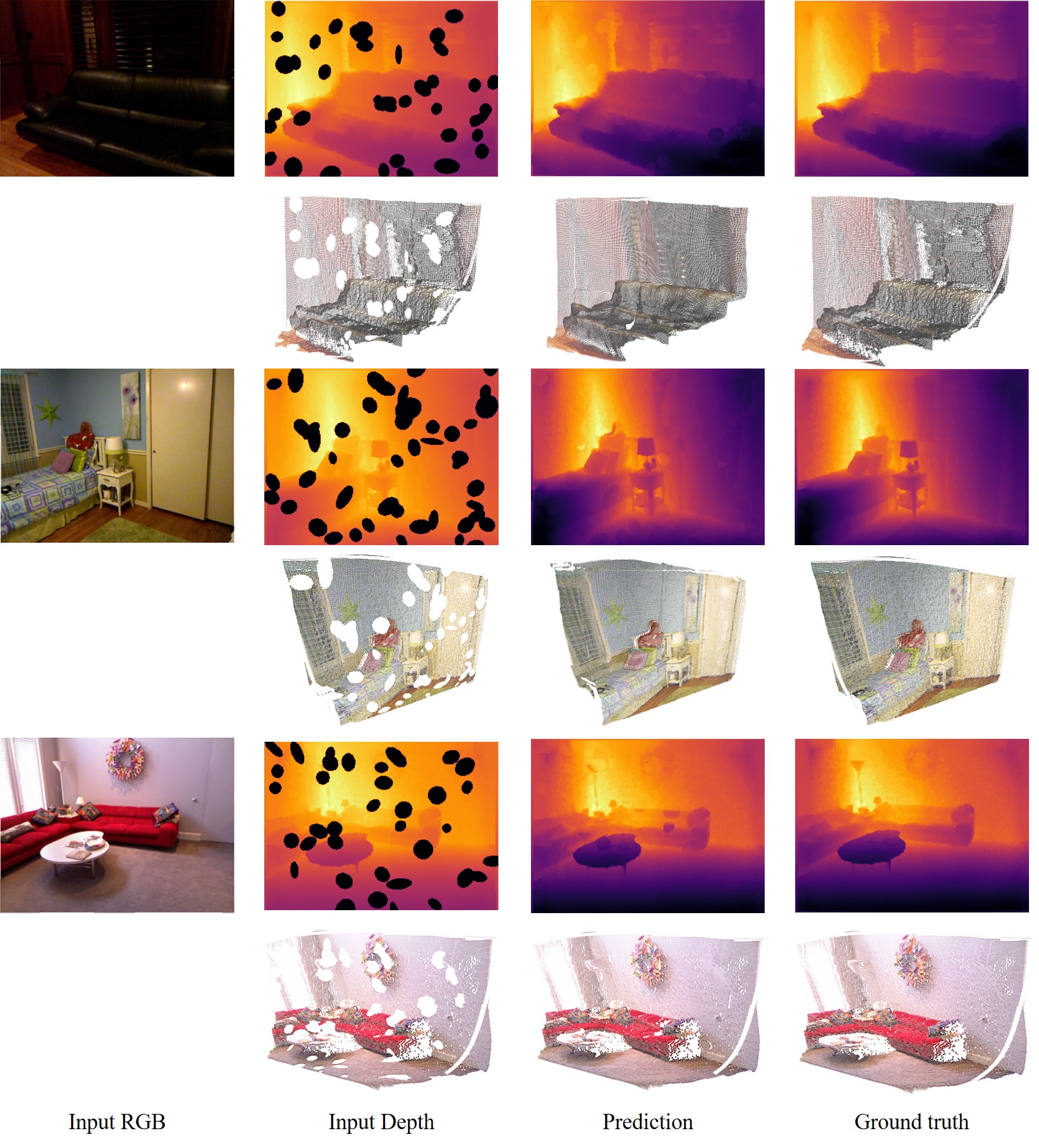}
\end{center}
\vspace{-3mm}
   \caption{Qualitative results on NYUV2 dataset. For every example, first row from left to right: input RGB, input depth, predicted depth, groundtruth depth; second row from left to right: input point cloud, predicted point cloud, groundtruth point cloud. Point clouds are rendered in a novel viewpoint. Please zoom in to see details.} 
\label{fig:nyu_results}
\end{figure*}

We have done experiments on the NYUV2 dataset~\cite{Silberman:ECCV12} to evaluate the performance of our method on general scenes and non-transparent objects. We corrupt the depth map by randomly creating some large holes. Our models are trained to predict the complete depth map given the corrupted depth map and RGB image. As shown in Figure~\ref{fig:nyu_results}, our method can predict reasonable missing depth in general scenes.

\section{Failure Cases}
\begin{figure*}
\begin{center}
   \includegraphics[width=1.0\linewidth]{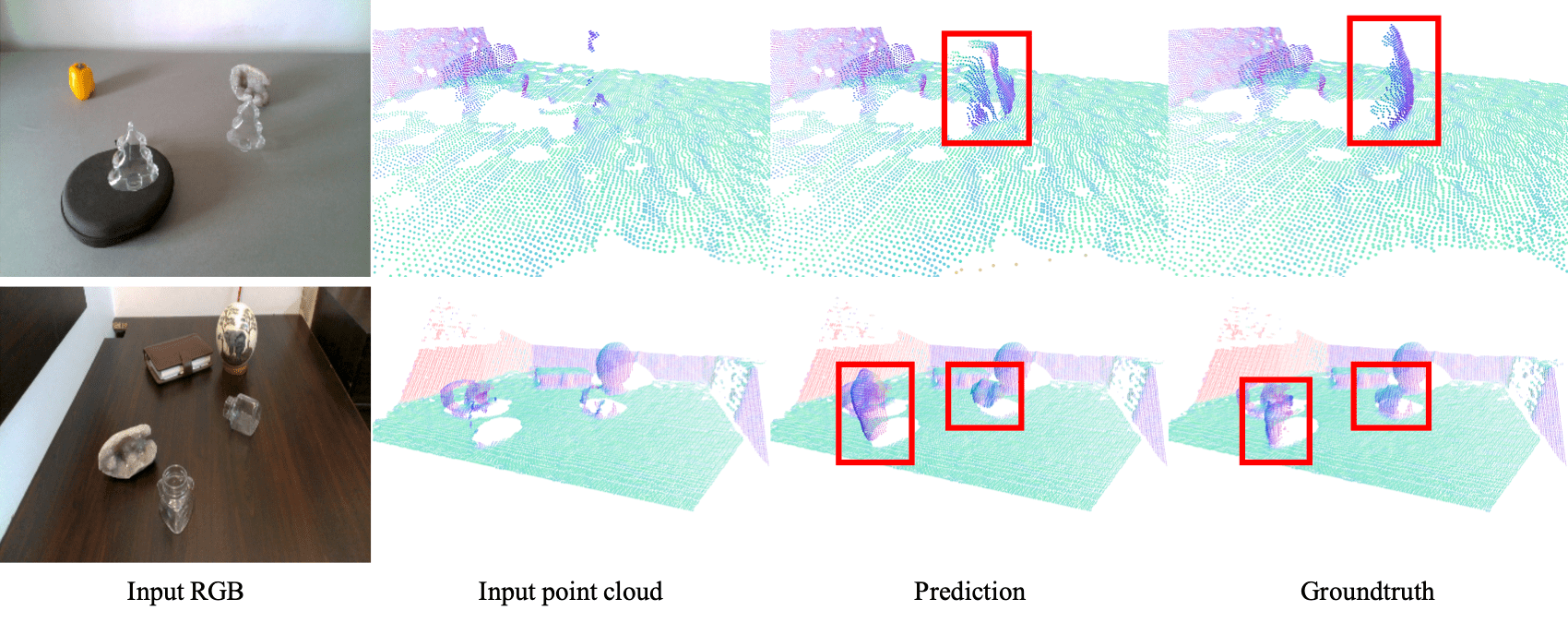}
\end{center}
\vspace{-3mm}
   \caption{Failure Cases. First row: pixels of the same object are classified into different terminating voxels, leading to a crack in the reconstructed object. Second row: there is no explicit constraint to force objects contacting the table, leading to objects floating in the air. Please zoom in to see details.} 
\label{fig:fail_case}
\end{figure*}

Figure~\ref{fig:fail_case} provides examples where our approach fails to complete depth of transparent objects from a single RGB-D image. The first limitation (first row) is that pixels of the same object may be classified into different terminating voxels, thus there might be a crack in the reconstructed object. The second limitation (second row) is that there is no explicit constraint in our approach to force objects contacting the table, leading to objects floating in the air.

\section{Discussion and Future Works}
There are several interesting directions for future works. We can extend our pipeline by treating each pixel's projection as a cone to account for the lateral noise. Generating training data with a realistic depth noise model~\cite{nguyen2012modeling} helps to improve the robustness of our method. We also plan to investigate depth completion of transparent objects in cluttered scenes with heavy occlusion.

\begin{table}[ht]
\setlength\tabcolsep{1.5pt}
\centering
\begin{tabular}{c|cccccc}
\toprule
Refinement & RMSE$\downarrow$ & REL$\downarrow$ & MAE$\downarrow$  & $\delta_{1.05}\uparrow$ & $\delta_{1.10}\uparrow$ & $\delta_{1.25}\uparrow$ \\
\hline
\multicolumn{1}{c|}{} & \multicolumn{6}{c}{ClearGrasp Syn-known} \\
\hline
$\times$ & 0.014 & \textbf{0.015} & \textbf{0.009} & 94.36 & 97.52 & 99.51 \\
\checkmark & \textbf{0.012} & 0.017 & \textbf{0.009} & \textbf{94.79} & \textbf{98.52} & \textbf{99.67} \\
\hline
\multicolumn{1}{c|}{} & \multicolumn{6}{c}{ClearGrasp Syn-novel} \\
\hline
$\times$ & 0.033 & 0.048 & 0.026 & 64.91 & 87.34 & \textbf{99.22} \\
\checkmark & \textbf{0.028} & \textbf{0.045} & \textbf{0.023} & \textbf{68.62} & \textbf{89.10} & 99.20 \\
\hline
\multicolumn{1}{c|}{} & \multicolumn{6}{c}{ClearGrasp Real-known} \\
\hline
$\times$ & \textbf{0.027} & \textbf{0.032} & \textbf{0.019} & \textbf{83.50} & \textbf{92.71} & 98.57 \\
\checkmark & 0.028 & 0.033 & 0.020 & 82.37 & 92.28 & \textbf{98.63} \\
\bottomrule
\end{tabular}
    \caption{Depth refinment model. $\times$ denotes prediction from first stage while \checkmark denotes prediction from refinement model.}
\label{table:refine_supp}
\end{table}

\begin{table}[ht]
\setlength\tabcolsep{1.5pt}
\centering
\begin{tabular}{cc|cccccc}
\toprule
RGB & Depth & RMSE$\downarrow$ & REL$\downarrow$ & MAE$\downarrow$  & $\delta_{1.05}\uparrow$ & $\delta_{1.10}\uparrow$ & $\delta_{1.25}\uparrow$ \\
\hline
\multicolumn{2}{c|}{} & \multicolumn{6}{c}{ClearGrasp Syn-known} \\
\hline
\checkmark & \checkmark &  \textbf{0.014} & \textbf{0.015} & \textbf{0.009} & \textbf{94.36} & \textbf{97.52} & \textbf{99.51} \\
& \checkmark & 0.061 & 0.093 & 0.050 & 46.53 & 72.16 & 92.15 \\
\checkmark & &  0.031 & 0.045 & 0.026 & 70.73 & 90.50 & 98.76 \\
\hline
\multicolumn{2}{c|}{} & \multicolumn{6}{c}{ClearGrasp Syn-novel} \\
\hline
\checkmark & \checkmark &  \textbf{0.033} & \textbf{0.048} & \textbf{0.026} & \textbf{64.91} & \textbf{87.34} & \textbf{99.22} \\
& \checkmark & 0.063 & 0.102 & 0.055 & 35.11 & 60.42 & 92.80 \\
\checkmark & &  0.075 & 0.119 & 0.066 & 34.70 & 54.81 & 84.16 \\
\hline
\multicolumn{2}{c|}{} & \multicolumn{6}{c}{ClearGrasp Real-known} \\
\hline
\checkmark & \checkmark &  \textbf{0.027} & \textbf{0.032} & \textbf{0.019} & \textbf{83.50} & \textbf{92.71} & \textbf{98.57} \\
& \checkmark & 0.071 & 0.098 & 0.055 & 38.30 & 67.51 & 91.53 \\
\checkmark & &  0.080 & 0.124 & 0.074 & 30.05 & 53.47 & 83.14 \\
\bottomrule
\end{tabular}
    \caption{Ablation studies for effect of different modalities}
\label{table:embeddings_supp}
\end{table}

\begin{table}[ht]
\setlength\tabcolsep{1.5pt}
\centering
\scalebox{0.8}{
\begin{tabular}{ccc|cccccc}
\toprule
Ray Info & Pos. Enc & \# Vox & RMSE$\downarrow$ & REL$\downarrow$ & MAE$\downarrow$  & $\delta_{1.05}\uparrow$ & $\delta_{1.10}\uparrow$ & $\delta_{1.25}\uparrow$ \\
\hline
\multicolumn{3}{c|}{} & \multicolumn{6}{c}{ClearGrasp Syn-known} \\
\hline
\checkmark & \checkmark & $8^3$ & 0.014 & \textbf{0.015} & \textbf{0.009} & \textbf{94.36} & 97.52 & 99.51 \\
 & N/A & $8^3$ & 0.034 & 0.032 & 0.019 & 86.15 & 93.60 & 97.92 \\
\checkmark & & $8^3$ & 0.018 & 0.024 & 0.013 & 89.41 & 97.33 & 99.60 \\
\checkmark & \checkmark & $4^3$ & \textbf{0.013} & 0.017 & \textbf{0.009} & 94.04 & \textbf{98.00} & \textbf{99.70} \\
\checkmark & \checkmark & $16^3$ & 0.017 & 0.021 & 0.011 & 90.62 & 97.06 & 99.45 \\
\hline
\multicolumn{3}{c|}{} & \multicolumn{6}{c}{ClearGrasp Syn-novel} \\
\hline
\checkmark & \checkmark & $8^3$ & 0.033 & \textbf{0.048} & 0.026 & \textbf{64.91} & 87.34 & \textbf{99.22} \\
 & N/A & $8^3$ & 0.066 & 0.089 & 0.050 & 49.73 & 70.88 & 91.30 \\
\checkmark & & $8^3$ & 0.041 & 0.057 & 0.031 & 58.88 & 82.36 & 97.73 \\
\checkmark & \checkmark & $4^3$ & \textbf{0.030} & 0.049 & \textbf{0.025} & 64.04 & \textbf{87.69} & 99.09 \\
\checkmark & \checkmark & $16^3$ & 0.040 & 0.057 & 0.032 & 61.11 & 83.85 & 97.60 \\
\hline
\multicolumn{3}{c|}{} & \multicolumn{6}{c}{ClearGrasp Real-known} \\
\hline
\checkmark & \checkmark & $8^3$ & \textbf{0.027} & \textbf{0.032} & \textbf{0.019} & \textbf{83.50} & \textbf{92.71} & \textbf{98.57} \\
 & N/A & $8^3$ & 0.066 & 0.072 & 0.043 & 61.64 & 77.98 & 91.99 \\
\checkmark & & $8^3$ & 0.032 & 0.039 & 0.024 & 78.07 & 90.81 & 96.93 \\
\checkmark & \checkmark & $4^3$ & 0.031 & 0.040 & 0.024 & 74.33 & 90.53 & 98.47 \\
\checkmark & \checkmark & $16^3$ & 0.035 & 0.044 & 0.025 & 71.04 & 83.90 & 97.80 \\
\bottomrule
\end{tabular}}
    \caption{Ablation Study for different design choices such as including ray information in the embedding, applying positional encoding, and the number of voxels on the accuracy}
\label{table:design_choices_supp}
\end{table}

\begin{table}[ht]
\setlength\tabcolsep{1.5pt}
\centering

\begin{tabular}{c|cccccc}
\toprule
Data & RMSE$\downarrow$ & REL$\downarrow$ & MAE$\downarrow$  & $\delta_{1.05}\uparrow$ & $\delta_{1.10}\uparrow$ & $\delta_{1.25}\uparrow$ \\
\hline
\multicolumn{1}{c|}{} & \multicolumn{6}{c}{ClearGrasp Syn-known} \\
\hline
CG+Omni & \textbf{0.014} & \textbf{0.015} & \textbf{0.009} & \textbf{94.36} & \textbf{97.52} & \textbf{99.51} \\
Omni & 0.063 & 0.106 & 0.053 & 41.80 & 65.80 & 89.98 \\
CG & 0.023 & 0.031 & 0.017 & 83.56 & 95.04 & 99.23 \\
\hline
\multicolumn{1}{c|}{} & \multicolumn{6}{c}{ClearGrasp Syn-novel} \\
\hline
CG+Omni & \textbf{0.033} & \textbf{0.048} & \textbf{0.026} & \textbf{64.91} & \textbf{87.34} & \textbf{99.22} \\
Omni & 0.062 & 0.108 & 0.053 & 30.11 & 57.81 & 91.14 \\
CG & 0.041 & 0.063 & 0.034 & 52.69 & 79.42 & 98.05 \\
\hline
\multicolumn{1}{c|}{} & \multicolumn{6}{c}{ClearGrasp Real-known} \\
\hline
CG+Omni & \textbf{0.027} & \textbf{0.032} & \textbf{0.019} & \textbf{83.50} & \textbf{92.71} & \textbf{98.57} \\
Omni & 0.059 & 0.082 & 0.048 & 43.24 & 70.20 & 93.74 \\
CG & 0.040 & 0.053 & 0.031 & 65.71 & 84.27 & 96.60 \\
\bottomrule
\end{tabular}
    \caption{Quantitative effect of training data.}
\label{table:training_data_supp}
\end{table}

\begin{table}[ht]
\setlength\tabcolsep{1.5pt}
\centering
\scalebox{0.95}{
\begin{tabular}{c|cccccc}
\toprule
Ray Pooling & RMSE$\downarrow$ & REL$\downarrow$ & MAE$\downarrow$  & $\delta_{1.05}\uparrow$ & $\delta_{1.10}\uparrow$ & $\delta_{1.25}\uparrow$ \\
\hline
\multicolumn{1}{c|}{} & \multicolumn{6}{c}{ClearGrasp Syn-known} \\
\hline
Argmax & \textbf{0.014} & \textbf{0.015} & \textbf{0.009} & \textbf{94.36} & \textbf{97.52} & \textbf{99.51} \\
WeightedSum & 0.018 & 0.028 & 0.014 & 85.23 & 95.78 & 99.42 \\
\hline
\multicolumn{1}{c|}{} & \multicolumn{6}{c}{ClearGrasp Syn-novel} \\
\hline
Argmax & \textbf{0.033} & \textbf{0.048} & \textbf{0.026} & \textbf{64.91} & \textbf{87.34} & \textbf{99.22} \\
WeightedSum & \textbf{0.033} & 0.051 & 0.027 & 63.25 & 86.14 & 98.84 \\
\hline
\multicolumn{1}{c|}{} & \multicolumn{6}{c}{ClearGrasp Real-known} \\
\hline
Argmax & \textbf{0.027} & \textbf{0.032} & \textbf{0.019} & \textbf{83.50} & \textbf{92.71} & \textbf{98.57} \\
WeightedSum & 0.030 & 0.037 & 0.022 & 78.27 & 91.53 & 97.83 \\
\bottomrule
\end{tabular}}
    \caption{Ablation study for different ray pooling strategies.}
\label{table:ray_pooling_supp}
\end{table}

\begin{table}[ht]
\setlength\tabcolsep{1.5pt}
\centering
\scalebox{0.95}{
\begin{tabular}{c|cccccc}
\toprule
Candidates & RMSE$\downarrow$ & REL$\downarrow$ & MAE$\downarrow$  & $\delta_{1.05}\uparrow$ & $\delta_{1.10}\uparrow$ & $\delta_{1.25}\uparrow$ \\
\hline
\multicolumn{1}{c|}{} & \multicolumn{6}{c}{ClearGrasp Syn-known} \\
\hline
Learned offset & \textbf{0.014} & \textbf{0.015} & \textbf{0.009} & \textbf{94.36} & \textbf{97.52} & \textbf{99.51} \\
Sample points & 0.019 & 0.024 & 0.013 & 88.84 & 95.68 & 98.88 \\
\hline
\multicolumn{1}{c|}{} & \multicolumn{6}{c}{ClearGrasp Syn-novel} \\
\hline
Learned offset & \textbf{0.033} & \textbf{0.048} & \textbf{0.026} & \textbf{64.91} & \textbf{87.34} & \textbf{99.22} \\
Sample points & 0.035 & 0.057 & 0.028 & 59.17 & 83.40 & 97.61 \\
\hline
\multicolumn{1}{c|}{} & \multicolumn{6}{c}{ClearGrasp Real-known} \\
\hline
Learned offset & \textbf{0.027} & \textbf{0.032} & \textbf{0.019} & \textbf{83.50} & \textbf{92.71} & 98.57 \\
Sample points & 0.033 & 0.041 & 0.024 & 73.79 & 89.22 & \textbf{98.70} \\
\bottomrule
\end{tabular}}
    \caption{Ablation study for candidate points selection.}
\label{table:cand_points_supp}
\end{table}

\end{document}